\documentclass[10pt,twocolumn,letterpaper]{article}
\usepackage[pagenumbers]{iccv} 

\definecolor{iccvblue}{rgb}{0.21,0.49,0.74}
\usepackage[pagebackref,breaklinks,colorlinks,allcolors=iccvblue]{hyperref}

\usepackage{url}

\usepackage[dvipsnames]{xcolor}

\usepackage{amsmath}
\usepackage{amssymb}

\usepackage{booktabs}
\usepackage{multirow}
\usepackage{colortbl}

\usepackage{algorithm} 
\usepackage{algorithmic} 

\usepackage{pifont}
\usepackage{marvosym}
\usepackage{wasysym}

\usepackage{color}

\usepackage{graphicx}
\usepackage{subcaption}
\usepackage{makecell}

\usepackage{microtype}
\usepackage{wasysym}

\def\paperID{***} %
\def\confName{ArXiV}
\def\confYear{2025}

\title{Cross-Modal Few-Shot Learning: a Generative Transfer Learning Framework}

\author{
    Zhengwei Yang$^{1,2,3,}$\textsuperscript{*},  
    Yuke Li$^{4,}$\textsuperscript{*},  
    Qiang Sun$^{5}$,  
    Basura Fernando$^{3}$,  
    Heng Huang$^{4}$,  
    Zheng Wang$^{1,2}$\textsuperscript{†} \\
    {\small $^{1}$National Engineering Research Center for Multimedia Software, Institute of Artificial Intelligence,} \\ 
    {\small  School of Computer Science, Wuhan University} \\
    {\small $^{2}$Hubei Key Laboratory of Multimedia and Network Communication Engineering} \\
    {\small $^{3}$Centre for Frontier AI Research, Agency for Science, Technology and Research, Singapore} \\
    {\small $^{4}$University of Maryland, College Park} \\
    {\small $^{5}$University of Toronto and MBZUAI} \\
    {\small \textsuperscript{*}Equal Contribution, \textsuperscript{†}Corresponding Author: wangzwhu@whu.edu.cn}
}

\begin{document}

\maketitle

\begin{abstract}
Most existing studies on few-shot learning focus on unimodal settings, where models are trained to generalize to unseen data using a limited amount of labeled examples from a single modality. However, real-world data are inherently multi-modal, and such unimodal approaches limit the practical applications of few-shot learning. To bridge this gap, this paper introduces the Cross-modal Few-Shot Learning (CFSL) task, which aims to recognize instances across multiple modalities while relying on scarce labeled data. 
This task presents unique challenges compared to classical few-shot learning arising from the distinct visual attributes and structural disparities inherent to each modality.
To tackle these challenges, we propose a Generative Transfer Learning (GTL) framework by simulating how humans abstract and generalize concepts.
Specifically, the GTL jointly estimates the latent shared concept across modalities and the in-modality disturbance through a generative structure. Establishing the relationship between latent concepts and visual content among abundant unimodal data enables GTL to effectively transfer knowledge from unimodal to novel multimodal data, as humans did.
Comprehensive experiments demonstrate that the GTL achieves state-of-the-art performance across seven multi-modal datasets across RGB-Sketch, RGB-Infrared, and RGB-Depth.

\end{abstract}

\section{Introduction}

Collecting large amounts of labeled data in real-world applications is often prohibitively expensive, time-consuming, or simply impractical~\citep{tharwat2023survey, sheng2024deep}. Few-shot learning (FSL) has emerged as a viable solution, enabling models to generalize effectively using only a handful of labeled examples~\citep{song2023meta,chen2019closer, luo2023closer, ke2024revisiting}. 
However, existing few-shot methods face significant challenges in handling the increasing prevalence of multi-modal data, such as multi-spectral images or multimedia content, which differs significantly from the extensive RGB data typically used in research. These multi-modal data collected from various types of sources and modalities, such as different sensors or imaging protocols, are becoming increasingly essential in applications like surveillance~\citep{wu2024single, hu2022m} and medical image analysis~\citep{jiang2023understanding, mok2024modality}. 
This highlights the need for more advanced FSL frameworks capable of leveraging the complementary information inherent in multi-modal data~\cite{luo2023closer, jiang2023mewl}.

\begin{figure}[t]
    \centering
    \begin{subfigure}{0.99\linewidth}
        \centering
        \captionsetup{font=footnotesize, labelfont=footnotesize}
        \includegraphics[width=0.99\linewidth]{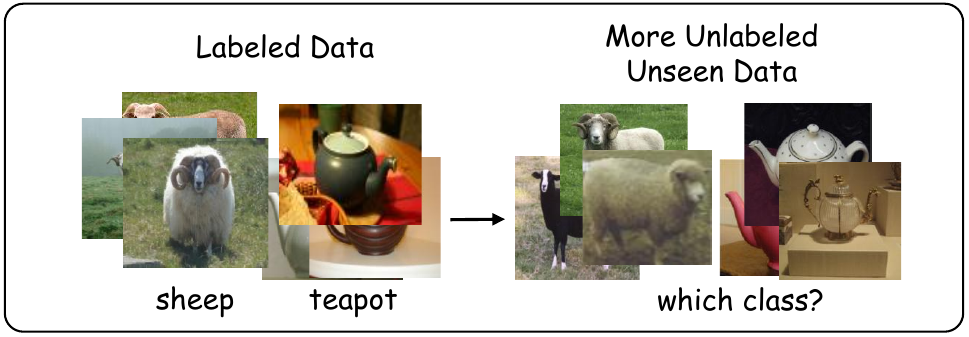}
        \caption{Classical supervised learning}
        \label{fig:motivation-a}
    \end{subfigure}
    \begin{subfigure}{0.99\linewidth}
        \centering
        \captionsetup{font=footnotesize, labelfont=footnotesize}
        \includegraphics[width=0.99\linewidth]{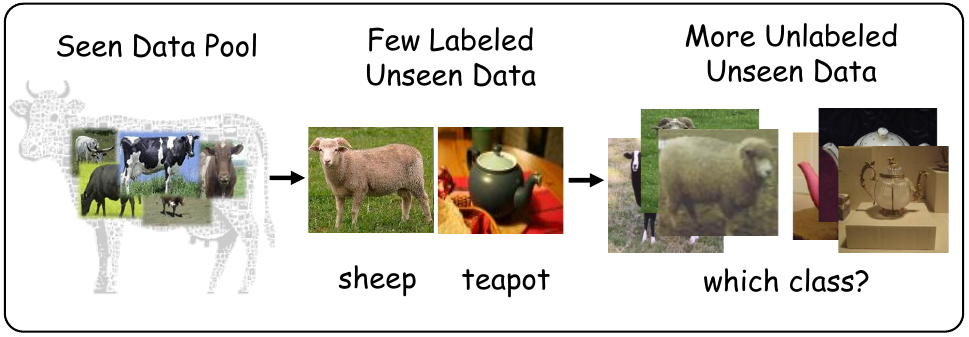}
        \caption{Classical few-shot learning}
        \label{fig:motivation-b}
    \end{subfigure}
    \begin{subfigure}{0.99\linewidth}
        \captionsetup{font=footnotesize, labelfont=footnotesize}
        \includegraphics[width=0.99\linewidth]{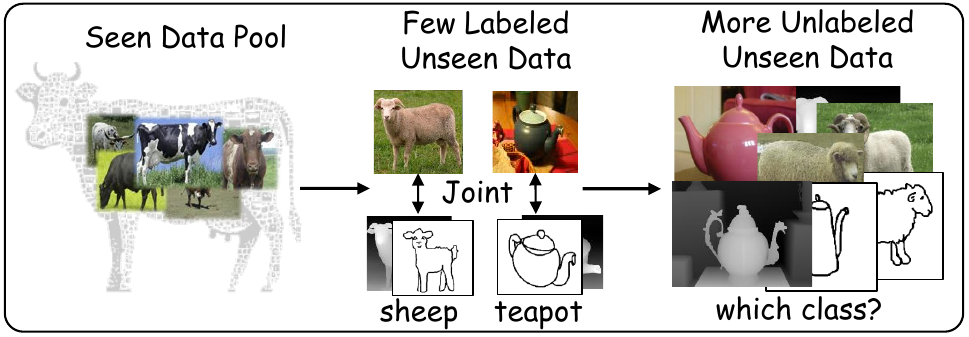}
        \caption{Cross-modal few-shot learning (ours)}
        \label{fig:motivation-c}
    \end{subfigure}
    \vspace{-1mm}
    \caption{\textbf{Comparison of recognition tasks.} (a) Classical recognition requires extensive labeled data within a single modality. (b) Few-shot recognition uses a few labeled samples in a single modality to classify unseen samples.
    (c) Our proposed CFSL involves few labeled multi-modal samples and aims to generalize to unseen multi-modal samples from the same classes, leveraging both seen and unseen data from different modalities.}
    \label{fig:motivation}
    \vspace{-5mm}
\end{figure}

Recent efforts have leveraged large pre-trained foundation models to extend their capabilities on novel multi-modal tasks, including tabular data~\citep{ye2024closerlookdeeplearning}, audio~\citep{Lin_2023_CVPR, duan2024cross}, and video classification~\citep{qing2023disentangling}, moving beyond traditional unimodal tasks like image classification~\citep{conti2023vocabulary}. Despite these advancements, the transfer of visual knowledge across different visual modalities remains relatively underexplored. Visual data, such as images and videos, constitute the most commonly studied data types; however, other visual modalities, such as infrared, depth, and sketches exhibit shared structural and contextual similarities with RGB data but possess unique attributes that make data collection and model adaptation more challenging.

In this case, this paper introduces a new Cross-modal Few-Shot Learning (CFSL) task designed to classify instances with multiple visual modalities with a limited number of labeled samples per class. 
In CFSL, the multi-modal data are organized into a support set and a query set. The support set comprises a few labeled examples from various modalities, while the query set contains unlabeled instances from corresponding classes and modalities that require classification.
As illustrated in~\Cref{fig:motivation}, unlike classical supervised learning and FSL confined to a single modality, CFSL engages data from multiple visual modalities. 
The primary challenge of CFSL arises from the inherent variability and domain gaps across different visual modalities, which complicates feature extraction and alignment.

\begin{figure}
    \centering
    \includegraphics[width=0.9\linewidth]{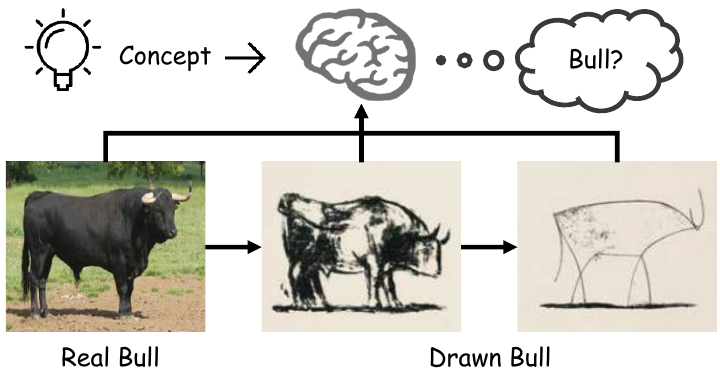}
    \caption{Illustration of the ability to generalize concepts like “Bull” across visual modalities (Pablo Picasso. \textit{The Bull}, 1945.).}
    \label{fig:motivation-d}
    \vspace{-3mm}
\end{figure}

Building on this understanding, previous studies~\citep{vinker2022clipasso, vinker2023clipascene, mukherjee2024seva} have
highlighted the crucial role of compact visual representations and underlying concepts in enabling humans to recognize target objects across different contexts, which can be leveraged to enhance CFSL.
As illustrated in~\Cref{fig:motivation-d}, the fundamental concepts like ``bull" can be readily learned with only a few examples, thanks to the humans' ability to generalize from limited data, regardless of visual modality~\citep{lake2020people, cao2021concept}. This capability, referred to as ``evocation", is grounded in the vast reservoir of previously encountered visual experiences that support recognition and generalization. 
Inspired by this observation, we hypothesize that models can achieve similar evocative capabilities by learning latent concepts from abundant unimodal data.

To this end, we propose a Generative Transfer Learning (GTL) framework to facilitate knowledge transfer between unimodal and multi-modal data. In the GTL framework, we posit that the latent concepts underlying target objects consist of two components: (1) the intrinsic concept that captures the core characteristic transferable across modalities, and (2) the in-modality disturbance that accounts for variations unique to each modality.
GTL aims to estimate these latent components and encode the relationships between the intrinsic concepts and visual content. 
Unlike existing methods that typically focus on modal adaptation~\cite{luo2023closer, ma2024cross} or invariant feature extraction~\cite{zhou2023revisiting, fu2023styleadv} which fail to jointly model both variant and invariant components within a unified generative framework, GTL explicitly model the invariant components to be effectively transferred across modalities and the variant components to be adapted to specific modality characteristics, much like human cognitive processes.

Within the GTL framework, our methodology consists of a two-stage training process. In the first stage, the {\bf generative learning stage}, the model learns latent concepts label-free, relying solely on pre-trained visual representations. This stage focuses on capturing the intrinsic concept and the variations in visual content across modalities, ensuring that the learned relationship between the latent concept and visual content is robust and transferable. In the second stage, the {\bf recognition stage}, the backbone network is frozen, and a separate classifier is trained on top of the learned latent intrinsic concepts to perform label prediction.

Our contributions are summarized as follows: 
\begin{itemize}
    \item  We introduce a new cross-modal few-shot learning task, which focuses on the connection and distinction between visual modalities, requiring models to perform recognition on multi-modal data with minimal labeled samples. This task better reflects real-world scenarios, where multi-modal data is scarce and diverse. 
    \item  We propose the generative transfer learning framework which explicitly models the variant and invariant components within a unified generative framework, enabling efficient knowledge adaption across modalities.
    \item We demonstrate the effectiveness of our method through extensive experiments on seven multi-modal datasets across RGB-Sketch, RGB-Infrared, and RGB-Depth.
\end{itemize}

\section{Task Settings}

In this section, we formally define the proposed CFSL task and highlight its differences from the previous FSL task. Additionally, we provide an overview of the CFSL task, emphasizing the challenges posed by handling multiple visual modalities with limited labeled data.

\begin{figure}[t]
    \centering
    \begin{subfigure}{0.49\linewidth}
        \centering
        \captionsetup{font=footnotesize, labelfont=footnotesize}
        \includegraphics[width=0.99\linewidth]{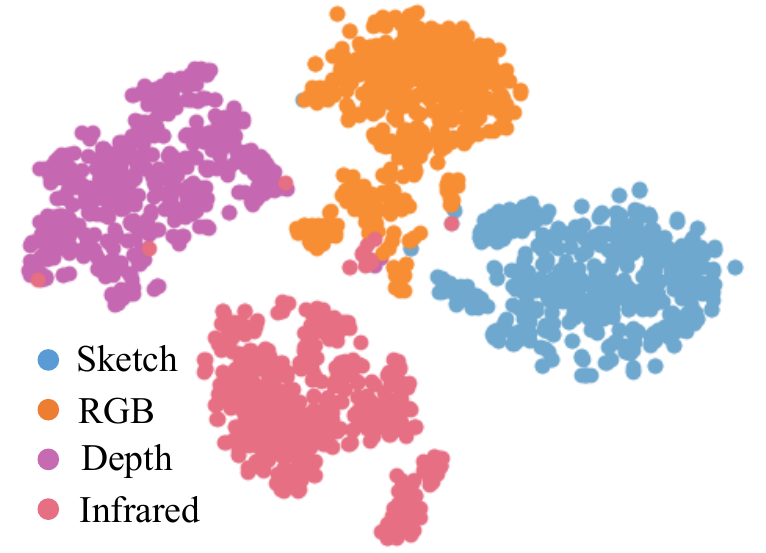}
        \caption{t-SNE of feature distribution \\ of different modalities}
        \label{fig:tsne2modality}
    \end{subfigure}
    \begin{subfigure}{0.49\linewidth}
        \centering
        \captionsetup{font=footnotesize, labelfont=footnotesize}
        \includegraphics[width=0.99\linewidth]{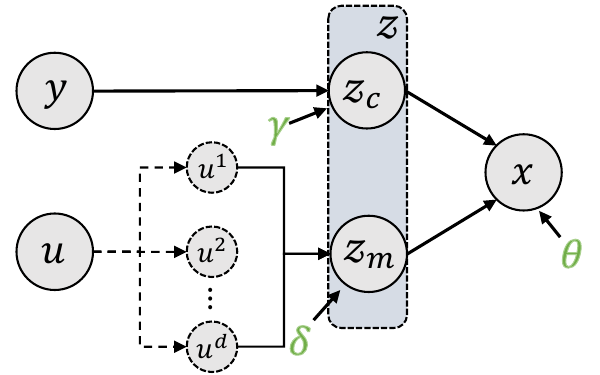}
        \caption{The data generating model \\ for the proposed CFSL}
        \label{fig:generative_graph}
    \end{subfigure}
    \vspace{-2mm}
    \caption{\textbf{The observation of the severe modality differences and the details of the proposed generative model.} (a) Illustration of the modality difference by the t-SNE clustering of the pre-trained CLIP~\citep{radford2021clip} features of different modalities. (b) The proposed generative process for the representation learning stage, the green symbols are assumed to be parameters that enable the models to adapt from base to novel data.}
    \label{fig:obervation&graph}
    \vspace{-0.5cm}
\end{figure}

\paragraph{Dataset Setup}
For the proposed CFSL task, the dataset $D = \{(x_m^i, y^i), y^i \in Y\}$ comprises a \textbf{base unimodal dataset} $D_\text{base} = \{(x_{m}^i, y^i), y^i \in Y_\text{base}, m=1\}$ and a \textbf{novel multimodal dataset} $D_\text{novel} = \{(x_m^i, y^i), m \in \{m^1, m^2, \dots, m^d \}, y^i \in Y_\text{novel}\}$. Here, $x_m^i$ denotes the feature vector of the $i$-th sample in modality $m$, $y^i$ is the corresponding class label. $Y_\text{base}$ and $Y_\text{novel}$ represent the set of class labels for the base and novel datasets, respectively. 
Notably, the class labels between the base and novel datasets are disjoint, aka, $Y_\text{base}\cap Y_\text{novel} = \varnothing$, such that $Y_\text{base}\cup Y_\text{novel} = Y$.

In classical FSL tasks, the goal is to train a model on $D_\text{base}$ with labels $Y_\text{base}$ and transfer this knowledge to improve the recognition performance on novel classes $Y_\text{novel}$ with the same modality, using only a few labeled examples for each class.
In contrast, our CFSL task introduces the added challenge of handling multi-modal data in the novel dataset $D_\text{novel}$ while the base dataset $D_\text{base}$ contains only unimodal data (e.g., RGB images). As shown in~\Cref{fig:tsne2modality}, there is a clear boundary among the multi-modal data.
This setting reflects real-world scenarios where models are typically trained on unimodal data but are expected to generalize effectively to multi-modal data.

The novel dataset $D_\text{novel}$ is further divided into:

\noindent $D_{\text{support}} = \{(x^i_m, y^i) | m\in\{m^1,m^2,\dots,m^d\}, y^i \in Y_\text{novel} \}$, a support set that contains a limited number of labeled samples per class from multiple modalities. Typically, this consists of $K$ labeled samples for each of $N$ classes. 

\noindent $D_{\text{query}} = \{x^j_m | m\in\{m^1,m^2,\dots,m^d\}\} $, a query set contains massive of unlabeled samples from the same classes as $D_{\text{support}}$, but potentially from the different modalities.

The fundamental challenge in this task is the limited labeled data in $D_{\text{support}}$, combined with the multi-modal data in $D_{\text{novel}}$. The scarcity of labeled data per class, along with the need to generalize across different modalities, makes this task significantly more challenging. 

As shown in \Cref{tab:vsmethod}, we provide a brief comparison between CFSL and existing tasks. For a more detailed discussion on related works and the distinctions between CFSL and existing tasks, please refer to Appendix A.

\begin{table}[t]
    \centering
    \footnotesize
    \setlength{\tabcolsep}{3pt}
    \begin{tabular}{c|c|c|c cccc}
    \toprule
     Data & Phrase & Properties & CFSL(ours) & FSL & CD-FSL & CM-FSL \\
    \midrule
     
     \multirow{2}[2]{*}[1ex]{Base} & \multirow{2}[2]{*}[1ex]{-} & RGB & $\checkmark$ & $\checkmark$ & $\checkmark$ & $\checkmark$  \\
       & & Other-Modal  & $\times$ & $\times$ & $\times$ & $\times$  \\
       
        \midrule
       
       \multirow{4}[2]{*}[0.1ex]{Novel} & \multirow{2}[2]{*}[1.1ex]{tune} & RGB & $\checkmark$ & $\checkmark$ & $\checkmark$/$\times$ & $\times$  \\
       & & Other-Modal & $\checkmark$ & $\times$ & $\times$/$\checkmark$ & $\checkmark$  \\
       \cline{2-7}
       & \multirow{2}[2]{*}[0.4ex]{test} & RGB & $\checkmark$ & $\checkmark$ & $\checkmark$/$\times$ & $\checkmark$  \\   
       & & Other-Modal & $\checkmark$ & $\times$ & $\times$/$\checkmark$ & $\times$  \\
         
    \bottomrule
    \end{tabular}
    \caption{\textbf{Task comparisons}. Unlike FSL~\cite{chen2019closer, luo2023closer}, CD-FSL (cross-domain)~\cite{guo2020broader, wang2020adversarial,xu2023deep}, and existing CM-FSL(cross-modality)~\cite{bhunia2022doodle}, CFSL uniquely integrates multi-modalities in the novel phase, requiring consistent handling of RGB and other modalities. 
    }
    \vspace{-3mm}
    \label{tab:vsmethod}
\end{table}

\paragraph{Task Overview}
To classify the samples in $ D_{\text{query}}$, the proposed CFSL involves a feature extraction function $ e_{\Phi} $, which is initially pre-trained on the large and fully labeled base dataset $D_{\text{base}}$. The goal is to adapt $ e_{\Phi} $ to $ e_{\Phi'} $, allowing it to extract discriminative features from all modalities in $ D_{\text{novel}} $ for classification. 
The adapted feature extractor $ e_{\Phi'} $ should ensure that samples from the same class $y^i$ are mapped close to each other in the feature space, enabling accurate classification regardless of the modality $m$. 
This adaptation process should help the classifier $ c_{\phi} $, parameterized by $ \phi $, in predicting the correct class labels for samples in $ D_{\text{query}} $ using features produced by $e_{\Phi'}$.
The adaptation allows the model to generalize across the novel modalities and classes, even with very few labeled examples in the support set.

\begin{figure*}
    \centering
    \includegraphics[width=0.9\linewidth]{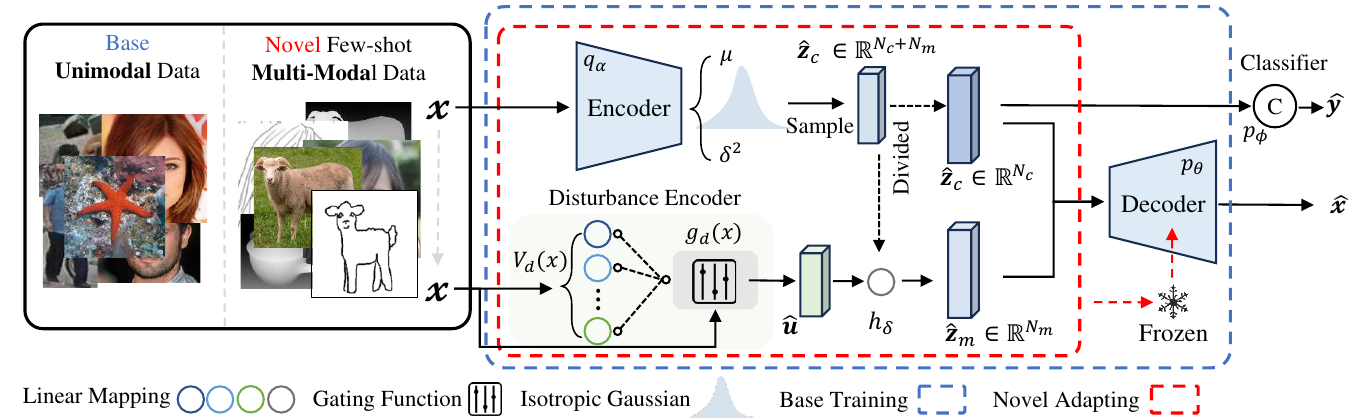}
    \vspace{-2mm}
    \caption{\textbf{The proposed GTL framework.} During the training on base data, all modules are trained (as in the \textcolor{blue}{blue} dashed box), but when adapting to novel data, the generator is frozen, and all other parts are tunable (as in the \textcolor{red}{red} dashed box). The classifier for recognition is separately initialized on the base and novel training since there is no overlap on class between them. }
    \label{fig:framework}
    \vspace{-3mm}
\end{figure*}

\section{Methodology}

This section formulates the problem, discusses network design, and introduces the learning objectives.

\subsection{Problem Formulation}

We assume that each observation $\mathbf{x}$ is generated from a nonlinear function $\mathbf{g}$:
\begin{align*}
\mathbf{x} 
= \mathbf{g}(\mathbf{z}) = \mathbf{g} (\mathbf{z}_m, \mathbf{z}_c),
\end{align*}
where $\mathbf{z}$ = ($\mathbf{z}_m$, $\mathbf{z}_c$), $\mathbf{z}_m$ contains in-modality disturbance, and $\mathbf{z}_c$ encapsulates latent intrinsic concept.

\Cref{fig:generative_graph} illustrates our data generating process. We formalize the probabilistic joint distribution of our data generating process by:
\begin{align}
p(\mathbf{x},\mathbf{z},\mathbf{u},y) =  p_{\delta}(\mathbf{z}_m | & \mathbf{u}^1,  \mathbf{u}^2, \dots, \mathbf{u}^d)  p_{\gamma}(\mathbf{z}_c | y) \notag \\ 
& p_{\theta}(\mathbf{x} | \mathbf{z}_m, \mathbf{z}_c)p(\mathbf{u})p(y). 
\end{align}

We use a VAE to model the generator $p_{\theta}(\mathbf{x} | \mathbf{z}_m, \mathbf{z}_c)$, where $\theta$ are the parameters, and $(\mathbf{z}_m, \mathbf{z}_c)$ are obtained by encoding $\mathbf{x}$ with parameters $\alpha$ via the posterior estimator $q_\alpha(\mathbf{z}|\mathbf{x})$.   The \(\delta\) and $\gamma$ are the parameters for modeling the distributions of $\mathbf{z}_m$ (via $p_{\delta}(\mathbf{z}_m | \mathbf{u}^1, \mathbf{u}^2, \dots, \mathbf{u}^d$)) and $\mathbf{z}_c$ (via $p_{\gamma}(\mathbf{z}_c | y)$), respectively.
To learn domain variable $ \mathbf{u} $ and predict the correct label $y$
, we also introduce two additional modules: a disturbance encoder $q_{\eta}({\mathbf{u}}|\mathbf{x})$ with parameters $\eta$, and a classifier $q_{\phi}(\hat{y}|\mathbf{z}_c)$ with parameters $\phi$. In whatfollowing, we will introduce each component in detail.

First, we introduce a posterior estimator parameterized by 
$\alpha$ as the encoder to learn the visual  latent representation $\mathbf{z}=\{\mathbf{z_c}, \mathbf{z_m}\}$ as:
\begin{equation} 
\mathbf{z} \sim q_{\alpha}(\mathbf{z} | \mathbf{x}).
\end{equation}

Second, since we lack supervision for learning the in-modality disturbance, we assume that this information can be estimated from the observed data $\mathbf{x}$. Therefore, a disturbance estimator, parameterized by $\eta$, is introduced to estimate the modality-relevant latent variable $\mathbf{u}$ as: 
\begin{equation}
    \mathbf{u} \sim q_{\eta}(\mathbf{u} | \mathbf{x}),
\end{equation}
where the multi-perspective domain variables $\mathbf{u}^1, \mathbf{u}^2, \dots, \mathbf{u}^d$ are generated from $\mathbf{u}$, capturing different perspectives of the modality.

Third, the modality-specific latent variable $\mathbf{z}_m$ is derived from the multi-perspective latent variables $\mathbf{u}^1, \mathbf{u}^2, \dots, \mathbf{u}^d$, governed by the learnable parameter $\delta$: 
\begin{equation}
    \mathbf{z}_m \sim p_{\delta}(\mathbf{z}_m | \mathbf{u}^1, \mathbf{u}^2, \dots, \mathbf{u}^d).
\end{equation}

Fourth, the latent intrinsic concept variable $\mathbf{z}_c$, which captures class-specific concept information, depends only on class label $y$ and is governed by the parameter $\gamma$:
\begin{equation}
 \mathbf{z}_c \sim p_{\gamma}(\mathbf{z}_c | y).
\end{equation}

Finally, the generator conceptualizes each observation $\mathbf{x}$ as being derived from a non-linear smooth mixing transformation, parameterized by $\theta$, involving latent variables $\mathbf{z}\subseteq\mathcal{Z}\in\mathbb{R}^n$, which are decomposed into two components $\mathbf{z}_c \in \mathbb{R}^{n_c}$ and $\mathbf{z}_m \in \mathbb{R}^{n_m}$, and 
\begin{equation}
    \mathbf{x} \sim p_{\theta}(\mathbf{x} | \mathbf{z}_m, \mathbf{z}_c).
\end{equation}

In the context of CFSL, the differing distributions of modality-relevant information $\mathbf{u}$ and class label $y$ between base and novel datasets necessitate adapting model parameters, including
the encoder $q_{\alpha}(\mathbf{z} | \mathbf{x})$, the disturbance estimator $q_{\eta}(\mathbf{u} | \mathbf{x})$, and the parameters responsible for generating the latent variables $\mathbf{z}_m$ and $\mathbf{z}_c$. However, the non-linear transformation, parameterized by $\theta$, remains invariant during the transfer learning stage, as it is assumed to capture the stable relationship between the latent concept and visual content. This invariance ensures that the model can adapt to novel data while preserving key generalizable components.


\subsection{Network Design}

In this section, we introduce the key components of our network, each playing a distinct role in modeling the cross-modal few-shot classification task. The specific details of the network architecture are presented in Appendix C. 
The operational framework of GTL is depicted in~\Cref{fig:framework}, detailing the training and adaptation process for CFSL scenarios.

\paragraph{Encoder}
We denote the encoder that performs posterior estimation as $q_{\alpha}(\mathbf{\hat{z}}|\mathbf{x})$, where $\mathbf{\hat{z}}$ represents the estimated latent variables. We assume that the two components $\mathbf{z}_m$ (modality-specific) and $\mathbf{z}_c$ (class-relevant) are conditionally independent given the observation $ \mathbf{x}$, allowing us to factorize the posterior distribution as:
\begin{equation}
    q_{\alpha}(\mathbf{z} | \mathbf{x}) = q_{\alpha}(\mathbf{z}_m | \mathbf{x}) q_{\alpha}(\mathbf{z}_c | \mathbf{x})
\end{equation}
Accordingly, we approximate the joint posterior distribution by assuming an isotropic Gaussian, characterized by a mean $\mu$ and covariance $\sigma^2$, as follows:
\begin{equation}
    q_{\alpha}(\mathbf{{z}}_c, \mathbf{{z}}_m | \mathbf{x}) \sim \mathcal{N}(\mu, \sigma^2),
\end{equation}
To learn this posterior distribution, we employ 1-layer MLPs with ReLU activation, a batch normalization layer, and a dropout layer as the estimator.

\paragraph{Disturbance Encoder}

The in-modality disturbance latent variable $\mathbf{z_m}$ is assumed to be transferable when adapting from the base to novel data. Since there is a lack of direct supervision regarding which representation corresponds to modality-specific information, we employ a flexible estimator to approximate the unobservable prior $p(\mathbf{z}_m)$. Inspired by the latent domain learning method~\citep{deecke2022visual}, we use a set of learnable gating functions $g(\mathbf{x})$ that assign each observation $\mathbf{x}$ to multiple latent domains, capturing different perspectives.

The estimated modality-relevant variable $\mathbf{\hat{u}}$ is used to guide the modality-specific variable $\mathbf{\hat{z}}_m$, and the estimation is given by:
\begin{equation}
\resizebox{0.9\columnwidth}{!}{$
\mathbf{\hat{z}}_m^{'} = h_{\delta, \mathbf{\hat{u}} \sim q_{\eta}}(\mathbf{\hat{u}}, \mathbf{\hat{z}}_m) \quad \text{and} \quad
q_{\eta}(\mathbf{\hat{u}} | \mathbf{x}) = \sum_{d=1}^{D} g_d(\mathbf{x}) V_d(\mathbf{x}) ,$}
\end{equation}
where $d$ is a hyperparameter that determines the number of latent domains, and each $V_d(\mathbf{x})$ is parameterized through a linear transformation. The function $h(\mathbf{u}, \mathbf{z_m})$ represents a linear aggregation of the domain variable $\mathbf{u}$ and the latent variables $\mathbf{z_m}$, with trainable parameter ${\delta}$.

\paragraph{Reconstruction}

The reconstruction module is responsible for generating an estimate of the observation $\mathbf{\hat{x}}$ based on the estimated latent variables $\mathbf{\hat{z}_c}$ (class-relevant information) and $\mathbf{\hat{z}_m^{'}}$ (modality-relevant information). We adopt a generator with a similar but reversed structure as the posterior estimator. While the posterior estimator encodes the latent variables, the generator decodes them back to the observed data.

The conditional distribution $p_\theta(\mathbf{x} | \mathbf{z}_c, \mathbf{z}_m^{'})$ is modeled by the generator, which consists of a 1-layer MLP with a ReLU activation function, a batch normalization layer, and a dropout layer. The reconstruction process is formulated as:
\begin{equation}
\resizebox{0.89\columnwidth}{!}{$
\hat{\mathbf{x}} = \text{Dropout} \left( \text{ReLU} \left( \text{MLP} ( \mathbf{\hat{z}{'}}) \right)\right) , \text{where} \quad \mathbf{\hat{z}}{'} = \{\mathbf{\hat{z}}_c, \mathbf{\hat{z}}_m^{'}\} .$}
\end{equation}

\paragraph{Classification}

The classifier $p_{\phi}(\mathbf{y} | \mathbf{z}_c)$, parameterized by $\phi$, can be implemented using a simple linear classifier. 
This involves a weight matrix $\mathbf{W} \in \mathbb{R}^{d \times c}$, where $d$ is the dimension of the latent variable $\mathbf{\hat{z}}_c$, and $c$ denotes the number of output classes. The classifier is trained by a standard cross-entropy loss function. 
The linear classifier computes the logits by linearly combining the latent variable $\mathbf{\hat{z}}_c$ with the weight matrix $\mathbf{W}$, followed by a softmax function to output class probabilities. The logits are computed as:
\begin{equation}
\mathbf{\hat{y}} = \text{softmax}(\mathbf{W}^\top \mathbf{\hat{z}}_c),
\end{equation}
where $\mathbf{\hat{y}}$ represents the predicted class probabilities.

\subsection{Learning Objectives}

\paragraph{Representation learning}

Based on the above generative learning structure,
the training objectives in the representation learning phase are formulated by the evidence lower bound (ELBO) as follows:
\begin{align}
    & \mathcal{L}_\text{ELBO}  =  \underbrace{\mathbb{E}_{q_{\mathbf{\hat{z}}_c, \mathbf{\hat{z}}_m^{'} | \mathbf{x}}} \left[ \ln{p_\theta(\mathbf{x} | \mathbf{\hat{z}}_c, \mathbf{\hat{z}}_m^{'})} \right]}_{\text{Reconstruction Loss}} \notag \\ 
 & - \lambda \underbrace{\mathbb{E}_{\mathbf{\hat{z}}_c, \mathbf{\hat{z}}_m^{'} \sim  q_{\alpha}, q_{\eta}, h_{\delta}} \left[ \log q(\mathbf{\hat{z}}_c, \mathbf{\hat{z}}_m^{'} | \mathbf{x}) - \log p(\mathbf{z}) \right]}_{\text{KL Divergence}},
\label{eq:elbo}
\end{align}
where the reconstruction term ensures that the model accurately reconstructs the input data from the latent variables $\mathbf{z}_c$ and $\mathbf{z}_m$, while the KL term regularizes the latent space to ensure the learned features align with the underlying data distribution. The hyperparameter $\lambda$ controls the trade-off between reconstruction accuracy and regularization strength.

\paragraph{Classification learning} 

In this phase, the cross-entropy loss measures the discrepancy between the predicted and actual class labels.
\begin{equation}
\mathcal{L}_{\text{CE}} = -\mathbb{E}_{\mathbf{\hat{y}}} (\mathbf{y} \log \mathbf{\hat{y}}).
\label{eq:crossentrophy}
\end{equation}

\subsection{Training and Inference}

This section outlines the workflow, from initial training on the base dataset to subsequent adaptation for novel data, clarifying the methodologies employed in each phase.

\paragraph{Phase 1: Training on Base Data}

The initial training phase is fundamental as it establishes the distinction between transferable and non-transferable components within the model. During this phase, all modules are trained using the base dataset $D_\text{base}$. Specifically, the posterior estimator ($\alpha$), the disturbance estimator and aggregator ($\eta$ and $\delta$), and the generator ($\theta$) are jointly trained using Eq.~\eqref{eq:elbo}. Afterward, the classifier ($\phi$) is trained using Eq.~\eqref{eq:crossentrophy}. The primary goal is to robustly encode domain-specific variations and content-specific features into separate latent spaces.

\paragraph{Phase 2: Transfer Learning for Adapting to Novel Data}

In this phase, the model is exposed to a novel data support set $D_\text{support}$. Given our assumption that the relationship between latent representations and visual content remains consistent across both base and novel datasets, we freeze the generator ($\theta$) in its trained state. This decision ensures that the foundational decoding process, which reconstructs visual content from latent representations, remains stable and unaffected by new data variability. We first fine-tune the posterior estimator ($\alpha$), the disturbance estimator, and aggregator ($\eta$ and $\delta$) using Eq.~\eqref{eq:elbo}, and then update the classifier ($\phi$) with a few labeled examples using Eq.~\eqref{eq:crossentrophy}. 

\begin{table*}[t]
    \centering
    \footnotesize
	\setlength{\tabcolsep}{14pt}
	\begin{tabular}{lccccc}
	\toprule
	\multirow{2}[2]{*}{Methods} & \multirow{2}[2]{*}{Venues} & {all-way-1-shot} & {all-way-5-shot} & {5-way-1-shot} &  {5-way-5-shot} \\ 
        \cmidrule(lr){3-4} \cmidrule(lr){5-6} & & \textit{Acc@avg}&\textit{Acc@avg}&\textit{Acc@avg}&\textit{Acc@avg} \\
	\midrule
        AGW~\cite{ye2021deep} & TPAMI'21 & 21.1 (23.9 / 18.2) & 46.0 (50.4 / 41.4) & 45.2 (47.0 / 43.2) & 71.7 (82.8 / 59.1)  \\
         Closer~\cite{luo2023closer} & ICML'23 & 27.6 (29.9 / 25.1) & 33.8 (18.1 / 50.6) & 28.8 (24.9 / 33.1) & 33.7 (16.8 / 52.4) \\
        Style~\cite{fu2023styleadv} & CVPR'23 & 32.9 (35.2 / 46.6) & 40.2 (40.3 / 40.9) & 46.6 (51.4 / 41.3) &  70.4 (78.1 / 61.9)  \\
        LDP-net~\cite{zhou2023revisiting} & CVPR'23 & 40.7 (39.7 / 41.7) & 44.1 (41.3 / 47.1) & 43.0 (49.8 / 46.5) & 66.7 (76.5 / 55.8) \\
         C2-Net~\cite{ma2024cross} & AAAI'24 & 40.2 (24.7 / 56.7) & 42.0 (26.5 / 58.6) & 44.2 (40.8 / 48.0) & 57.8 (50.1 / 66.3) \\

        TransReID~\cite{he2021transreid} & ICCV'21 & 46.0 (23.4 / 70.2) & 64.3 (59.9 / 67.0) & 78.6 (64.6 / 94.0) & 89.6 (83.4 / 96.4)\\
        CLIP-ReID~\cite{li2023clip} & AAAI'23 & 60.7 (56.2 / 65.4) & 81.4 (77.9 / 85.0) & 83.2 (76.9 / 90.1)  &  94.0 (95.3 / 92.5) \\
        \midrule

        Ours& - &  \bf{63.8 (58.8 / 69.1)} & \bf{82.9 (79.8 / 86.3)} & \bf{84.5 (81.3 / 87.9)} & \bf{94.1 (95.4 / 92.9)}  \\
        \bottomrule
	\end{tabular}
        \vspace{-2mm}
        \caption{Quantitative results of ours and other sota competitors on \textsc{SKETCHY} dataset. The best results are marked as \textbf{BOLD.} The A(B/C) metrics under the \textit{Acc@avg} stands for the average Top-1 accuracy for the A: mixed-modality, B: sketch, and C: RGB data. }
	\vspace{-3mm}
        \label{tab:sketchy}
\end{table*}

\begin{table*}[t]
    \centering
    \footnotesize
	\setlength{\tabcolsep}{14pt}
	\begin{tabular}{lccccc}
	\toprule
	\multirow{2}[2]{*}{Methods} & \multirow{2}[2]{*}{Venues} & {all-way-1-shot} & {all-way-5-shot} & {5-way-1-shot} &  {5-way-5-shot} \\ 
        \cmidrule(lr){3-4} \cmidrule(lr){5-6} & & \textit{Acc@avg}&\textit{Acc@avg}&\textit{Acc@avg}&\textit{Acc@avg} \\
	\midrule
        AGW~\cite{ye2021deep} & TPAMI'21 & 14.7 (32.0 / 13.0) & 40.1 (64.2 / 37.9) & 69.8 (79.5 / 68.8) & 75.9 (89.6 / 74.5) \\
        Closer~\cite{luo2023closer} & ICML'23 & 22.9 (34.3 / 21.8) & 48.8 (31.5 / 50.4) & 35.5 (47.4 / 34.2) & 50.9 (48.0 / 51.2) \\
        Style~\cite{fu2023styleadv} & CVPR'23 & 35.5 (37.1 / 35.3)  & 49.9(50.8 / 49.8) & 70.3 (69.9 / 70.3) & 76.2 (81.3 / 75.6) \\
        LDP-net~\cite{zhou2023revisiting} & CVPR'23 & 38.7 (39.8 / 38.5) & 47.2(52.6 / 46.7) & 70.1 (69.7 / 70.2) & 74.0 (80.2 / 73.5) \\
        C2-Net~\cite{ma2024cross} & AAAI'24 & 39.1 (28.8 / 40.1) & 50.1 (46.9 / 50.4) & 60.1 (60.9 / 70.0) & 65.3 (74.9 / 74.3) \\
        TransReID~\cite{he2021transreid} & ICCV'21 & 38.0 (29.9 / 38.8) & 61.1 (68.9 / 60.3) & 83.5 (63.3 / 85.7) & 93.3 (92.8 / 93.4) \\
        CLIP-ReID~\cite{li2023clip} & AAAI'23 & 46.6 (44.2 / 46.8) & 74.7 (75.8 / 74.6) & 91.5 (79.7 / 92.8)  &  97.1 (94.9 / 97.3) \\
        \midrule
        Ours & - & \bf{47.1 (46.3 / 47.3)} & \bf{74.8 (76.1 / 74.7)} & \bf{92.2 (85.1 / 93.0)} & \bf{98.0 (98.1 / 98.0)} \\
        \bottomrule
	\end{tabular}
        \vspace{-2mm}
        \caption{Quantitative results of ours and other sota competitors on \textsc{TU-Berlin} dataset.}
	\label{tab:tuberlin}
	\vspace{-4mm}
\end{table*}

\paragraph{Inference with Novel Data Query Set}

In the inference phase, samples from the novel data query set $D_\text{query}$ are first processed by our posterior estimator ($\alpha$), which generates latent intrinsic concepts based on the learned model. These representations are then used by the trained classifier ($\phi$) to make predictions. We select the prediction $\hat{y}$ with the maximum value, which corresponds to the highest predicted probability across all possible classes, to determine the most likely class label.

\section{Experiments}

This section covers experimental setups and benchmark results. The setups detail datasets, split strategy, evaluation protocol, and implementation details, while the results present outcomes, ablation studies, and hypothesis verifications.

\subsection{Experimental Setups}

\paragraph{Datasets}

We conduct extensive experiments on seven multi-modal datasets: four RGB-sketch, including \textsc{Sketchy}~\citep{sangkloy2016sketchy} (75K sketches, 73K RGB, 125 classes),  \textsc{TU-Berlin}~\citep{sketch12, zhang2016sketchnet} (20K sketches, 204K RGB, 250 classes), \textsc{Mask1K}~\citep{lin2023beyond, zheng2015scalable} (4.7K sketches, 996 ids, 6 styles), and  \textsc{SKSF-A}~\citep{yun2024stylized} (938 pairs, 134 ids, 7 styles); one RGB-infrared \textsc{SYSUMM01}~\citep{wu2017rgb} (22K RGB, 16K infrared, 491 ids); and two RGB-depth, which includes \textsc{RobotPKU}~\citep{liu2017online} (16K pairs, 180 ids) and \textsc{Washington RGB-D}~\citep{lai2011large} (250K images, 51 classes).

\paragraph{Dataset Split} 

For experiments on \textsc{Sketchy}, we follow the dataset splitting scheme used by~\citet{bhunia2022doodle}, which contains 64 base classes, while the remaining 61 classes are used as the novel set to make the task more challenging. 
For \textsc{TU-Berlin}, we split into 125 base and 125 novel classes following~\citet{bandyopadhyay2024generalised}. 
Similarly, for \textsc{Washington RGB-D}, we split into 25 base and 26 novel classes.
For the biometric multi-modal dataset, all data in \textsc{Mask1K}, \textsc{SKSF-A}, \textsc{SYSUMM01}, \textsc{RobotPKU} are considered novel;
we use commonly adopted person and face datasets MSMT17~\citep{wei2018person} and CelebA-HQ~\citep{Tero18CelebHQ} as base sets for person and face recognition, respectively. Each base set is further divided into training, validation, and testing subsets (60\%:20\%:20\%) following \citet{bhunia2022doodle}.

\paragraph{Evaluation Protocol} 

We report the results under two settings: {\bf all-way-$k$-shot}~\citep{ju2022prompting, lilearning} and {\bf standard 5-way-$k$-shot}~\citep{luo2023closer}. In the all-way-$k$-shot setting, all classes are presented to the model, with $k$ examples randomly sampled per class to form the support set $S$, while the remaining examples constitute the query set $Q$. In the 5-way-$k$-shot setting, 5 classes are randomly selected per episode, and for each of these 5 classes, $k$ examples are sampled to form the support set $S$, with the remaining examples used for the query set $Q$. 

In both settings, $k$ denotes the number of labeled samples per class in the support set $S$. After splitting the data into a base set and a novel set comprising $S \cup Q$, $k$ determines how many samples per class are included in $S$, with the remaining samples forming $Q$.

\textbf{Evaluation metric} 
We compare model performance using {\bf Top-1 Accuracy} over the novel query set, referred to as \textit{Acc@avg}, which measures the proportion of instances where the model's top predicted label matches the true label across mixed modalities. Additionally, we report the average Top-1 accuracy within each modality.

\paragraph{Implementation Details}

We implement our framework in PyTorch and run experiments on an NVIDIA RTX 2080Ti GPU. ViT-B/16~\citep{vit2020image}, pre-trained on the base data with CLIP~\citep{radford2021clip} for backbone initialization as the visual encoder to extract the visual representation $\mathbf{x}$. The hyperparameter $d$ is set to 128, and $\lambda$ is set to 1. We use the Adam optimizer~\cite{KingmaB14} for 60 epochs with an initial learning rate of $1e^{-3}$ during the representation learning stage and $1e^{-4}$ during the classification stage. The learning rate decreases to 10\% of the original value after 30 epochs, and weight decay is fixed at $1e{-4}$ for all settings.

\begin{figure*}[t]
    \centering
    \includegraphics[width=0.8\linewidth]{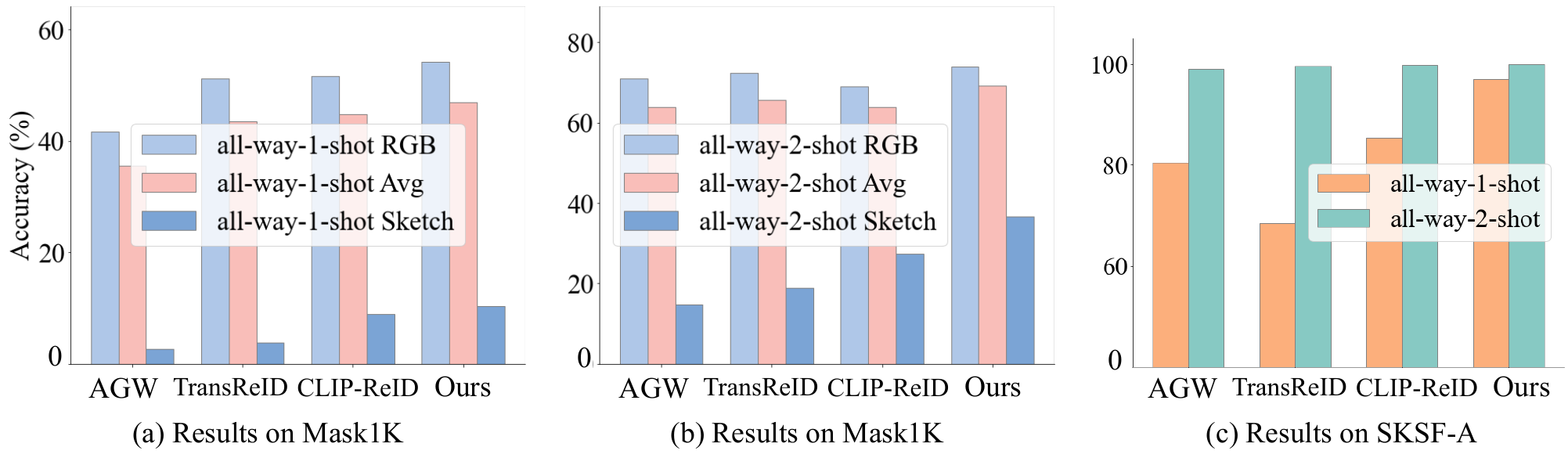}
    \vspace{-2mm}
    \caption{Experimental results of different shots on testing performance on (a) and (b) \textsc{Mask1k}, and (c) \textsc{SKSF-A}  datasets.}
    \label{fig:biometric_exp}
    \vspace{-3mm}
\end{figure*}

\subsection{Benchmark Results}

We benchmark our method against several FSL~\citep{ma2024cross,luo2023closer}, CD-FSL~\citep{fu2023styleadv, zhou2023revisiting} and fine-grained retrieval models~\citep{ye2021deep,he2021transreid, li2023clip}.

\paragraph{Comparison on  multi-modal category datasets} \quad

\textit{Sketchy}: ~\Cref{tab:sketchy} presents our method consistently outperforms all five leading benchmarks, with the highest improvement reaching 4.4\% in accuracy across different settings. Notably, in the aw1s setting, our approach achieves a 3.1\% improvement in overall accuracy compared to the second-best method. Additionally, our model shows gains in both individual modality classification metrics (sketch and RGB) across all settings.

\textit{TU-Berlin}: \Cref{tab:tuberlin} shows despite its larger scale and more abstract samples, GTL achieves the best performance across all accuracy metrics. 
In the 5-way-$k$-shot setting, it improves sketch modality accuracy by 5.4\%, highlighting its robustness in cross-modal few-shot recognition.

\begin{table}[t]
    \centering
    \footnotesize
    \setlength{\tabcolsep}{3pt}
    \begin{tabular}{l|cccc|cccc}
    \toprule
     \multirow{2}[2]{*}[1ex]{Method} &  \multicolumn{4}{c}{\textsc{Washington RGBD}} & \multicolumn{4}{c}{\textsc{RobotPKU}}  \\
     \cmidrule(lr){2-5}  \cmidrule(lr){6-9}
     & aw1s & aw5s &5w1s & 5w5s & aw1s & aw5s &5w1s & 5w5s \\
    \midrule
     Closer~\cite{luo2023closer} & 25.7 & 46.7 & 57.0 & 68.5 & 27.5 & 51.5 & 54.3 & 66.0\\
     Style~\cite{fu2023styleadv} & 32.5 & 53.3 & 57.8 & 73.9 & 38.4 & 60.9 & 57.7 & 75.7\\
     LDP-net~\cite{zhou2023revisiting} & 33.6 &  50.3 & 60.4 & 76.2 & 37.2 & 59.1 & 60.0 & 74.6\\
     CLIP-ReID~\cite{li2023clip} & 39.0 &  61.2 & 76.5 & 81.3 & 45.4 & 76.3 & 81.0 & 88.2\\
     GTL(ours) & \textbf{44.4} & \textbf{65.3} & \textbf{80.7} & \textbf{87.6} & \textbf{56.5} & \textbf{82.1} & \textbf{85.9} & \textbf{90.7}  \\
    \bottomrule
    \end{tabular}
    \caption{Quantitative results under \textit{Acc@avg} of ours and other sota competitors on \textsc{Washington RGBD} and \textsc{RobotPKU} datasets. ``aw1s, aw5s, 5w1s, 5w5s" is short for ``all-way-1-shot, all-way-5-shot, 5-way-1-shot, 5-way-5-shot", respectively.}
    \vspace{-4mm}
    \label{tab:vsDepth}
\end{table}

\textit{Washington RGB-D:} The left of~\Cref{tab:vsDepth} supports that the proposed GTL consistently outperforms existing methods, achieving 87.6\% accuracy in the 5w5s setting. This demonstrates its strong adaptability in RGB-depth scenarios.

\paragraph{Comparison on biometric multi-modal datasets} \quad

\textit{RobotPKU:} The right of~\Cref{tab:vsDepth} demonstrates GTL achieves sota compared to existing methods and achieves 90.7\% in 5w5s, showing its effectiveness in handling cross-modal identity recognition in dynamic environments. 
The higher performance on both depth datasets is attributed to their limited scene variation, making cross-modal alignment more effective.

\begin{table}[t]
    \centering
    \footnotesize
	\setlength{\tabcolsep}{8pt}
	\begin{tabular}{lcccc}
	\toprule
	\multirow{2}[2]{*}{Methods} & \multicolumn{4}{c}{\textsc{SYSUMM01}} \\ 
        \cmidrule(lr){2-5} & {aw1s} & {aw5s} & {5w1s} &  {5w1s} \\
	\midrule

        AGW~\cite{ye2021deep} & 60.5  & 73.2  & 63.8  & 77.9 \\
        TransReID~\cite{he2021transreid}  & 59.2 & 74.3 & 66.5 & 83.3  \\
        CLIP-ReID~\cite{li2023clip} &  68.5 & 81.7  & 70.4 & 87.4  \\
        \midrule
        Ours  & \bf{72.1} & \bf{86.9} & \bf{75.7} & \bf{90.0} \\
        \bottomrule
	\end{tabular}
        \caption{Comparing ours with other competitors on \textsc{SYSUMM01}. }
        \vspace{-3mm}
	\label{tab:supp_sysu}
\end{table}

\textit{SYSUMM01:} As shown in~\Cref{tab:supp_sysu}, our method achieves 72.1\% in the aw1s setting and 86.9\% in the aw5s setting. For 5-way settings, it reaches 75.7\% in 5w1s and 90.0\% in 5w5s, demonstrating strong performance across different retrieval configurations.

Given the limited modality data per class (8 samples for both \textsc{Mask1K} and \textsc{SKSF-A}), we decrease the $k$ to 1 and 2 and evaluate the models only under the all-way settings.

\textit{Mask1K:} The high number of classes and limited training samples introduce challenges for recognition. As shown in \Cref{fig:biometric_exp}a and \ref{fig:biometric_exp}b, our method achieves accuracy improvements of 2.2\% (k=1) and 5.4\% (k=2), with notable gains in individual modality classification metrics.

\textit{SKSF-A:} 
Since only one RGB image per class is available for fine-tuning, we report only sketch modality accuracy. As illustrated in \Cref{fig:biometric_exp}c, all methods achieve relatively high accuracy, indicating clear inter-class separability. Our method improves accuracy by 11.5\% for $k=1$, demonstrating strong adaptability to limited data.

\begin{table}
\centering
\footnotesize
\setlength{\tabcolsep}{10pt}
\begin{tabular}{lcc}
    \toprule
    & 
    all-way-1-shot & all-way-5-shot \\
    \cmidrule(lr){2-3} &\textit{Acc@avg}&\textit{Acc@avg}\\
    \midrule
      w/o $\mathbf{z}$ & 48.7 (40.0 / 58.0) & 67.8 (59.2 / 76.9) \\
      w/o $\mathbf{z}_m$ & 53.9 (41.7 / 67.0)  &  73.8 (63.9 / 84.3) \\
     GTL & \bf{63.8 (58.8 / 69.1)}  & \bf{82.9 (79.8 / 86.3)} \\
     \midrule
   $\text{GTL}_{T}$ & 44.2 (40.7 / 48.0) & 69.9 (65.9 / 74.0) \\
   $\text{GTL}_{FT}$ & 61.5 (56.6 / 66.8) & 81.1 (78.1 / 84.4) \\
   GTL & \bf{63.8 (58.8 / 69.1)}  & \bf{82.9 (79.8 / 86.3)} \\
    \bottomrule
\end{tabular} 
\vspace{-2mm}
\caption{Ablation studies of each component in GTL on \textsc{Sketchy}.}
\vspace{-3mm}
\label{tab:abalation_componets}
\end{table}

\subsection{Ablation Studies}

To verify the effectiveness of each module in GTL framework,~\Cref{tab:abalation_componets} presents the ablation study results. The first two rows show the results of removing components: ``w/o $\mathbf{z}$" excludes all latent variables, using only the classifier ($\phi$) for label prediction; ``w/o $\mathbf{z_m}$" removes only the disturbance latent variable $\mathbf{z_m}$. The lower rows compare training strategies. $\text{GTL}_{T}$ trains all modules from scratch, while $\text{GTL}_{FT}$ fine-tunes the generator during adaptation instead of keeping it fixed.
The last row represents our complete GTL framework, achieving the best results in both settings, as indicated by the bolded accuracy scores.

Comparing the results in~\Cref{tab:abalation_componets}, incorporating latent concept learning significantly improves performance. Including $\mathbf{z_m}$ (third row) increases RGB accuracy by 18.8\% and sketch accuracy by 11.1\% over the baseline (first row). Excluding modality-specific variables (``w/o $\mathbf{z_m}$", second row) improves the accuracy of RGB and sketch by 9\% and 1.7\%. Fixing the generator after pre-training (final row) results in the highest average accuracy of 19.6\%, outperforming the variant where the generator is fine-tuned.
Additional experiments on the hyperparameter selection of the learnable latent domain number $d$ are provided in the Appendix B.

\begin{figure}[t]
    \centering \includegraphics[width=0.9\linewidth]{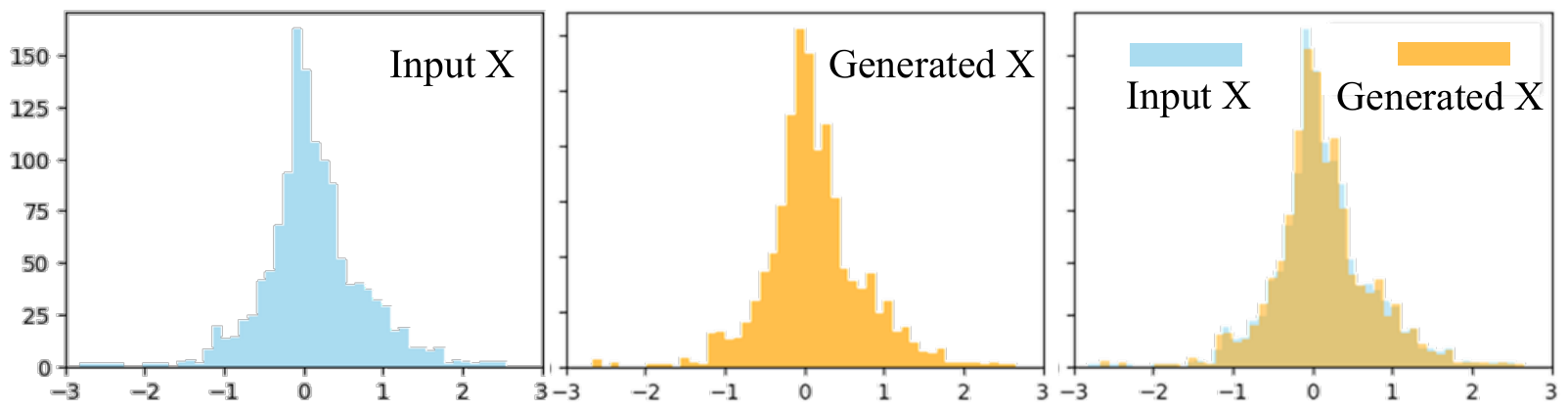}
     \vspace{-3mm}
    \caption{Distribution analysis among input and generated data.}
    \label{fig:distribution}
    \vspace{-4mm}
\end{figure}

\subsection{Discussions}

Beyond the ablation studies, this section examines our hypothesis from data distribution perspectives.

\paragraph{Assumption validation: Estimating generative structure}
To evaluate the generative structure, we analyze the distributions of input data, decoder-generated output in~\Cref{fig:distribution}. The left and middle histograms represent the distributions of the original input and the generated output, respectively, while the right plot overlays both distributions. The high similarity between the input and generated data suggests that the decoder effectively reconstructs the original distribution, supporting the validity of our generative assumption.

\paragraph{Assumption validation: Estimating in-modality disturbance}

\begin{figure}[t]
    \centering
    \includegraphics[width=0.98\linewidth]{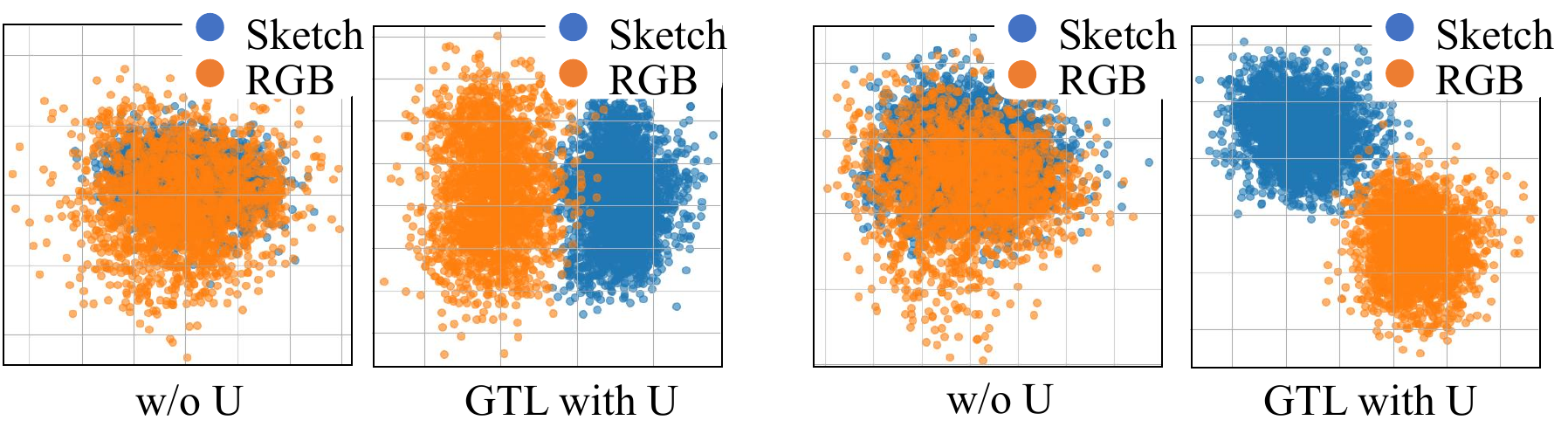}
    \vspace{-2mm}
    \caption{The t-SNE of the estimated in-modality disturbance representation on \textsc{Sketchy} (left) and \textsc{Mask1k} (right) datasets.}
    \label{fig:tsne_u}
    \vspace{-3mm}
\end{figure}

To examine the role of in-modality disturbances, we conduct t-SNE clustering on latent representations  $\mathbf{\hat{u}}$ from models trained with and without a disturbance estimator. As shown in~\Cref{fig:tsne_u}, without the estimator, latent representations from different modalities (e.g., RGB and sketch) are mixed, indicating the model struggles to distinguish modality-specific features. In contrast, with the estimator, modality-specific clusters emerge, demonstrating its effectiveness in preserving unique characteristics and separating in-modality disturbances.

\begin{figure}[t]
    \centering
    \includegraphics[width=0.99\linewidth]{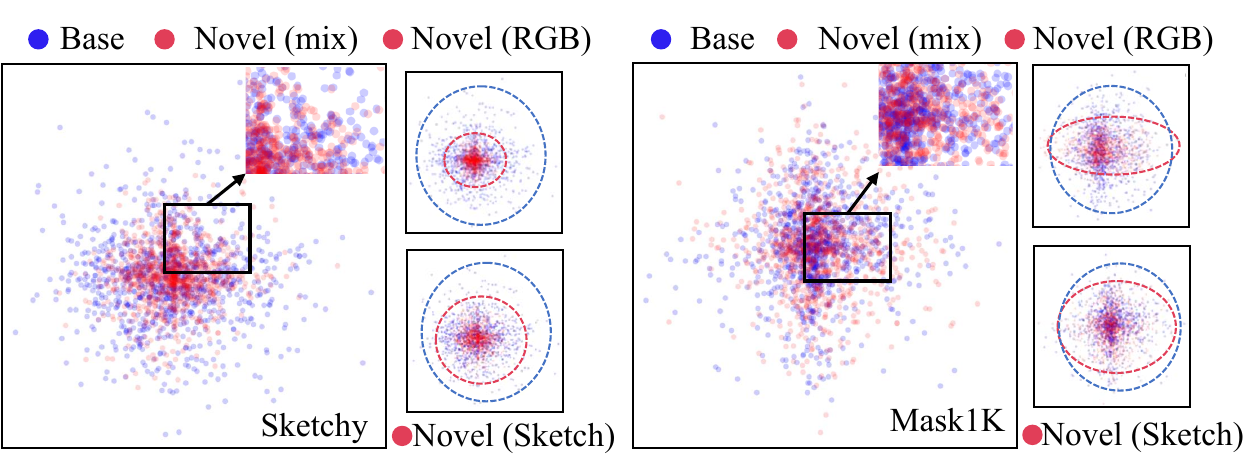}
     \vspace{-2mm}
    \caption{Visualizations of the pair plot of the learned latent representation and the visual representation on \textsc{Sketchy} (left) and \textsc{Mask1k} (right) datasets. \textcolor{blue}{Blue} and \textcolor{red}{red}  stands for model trained on based and novel data, respectively. The dashed circle with the corresponding color denotes the approximate scope of distributions. The (mix) denotes the trained data are multi-modal, and (RGB) and (Sketch) denote the unimodal data are used for training. }
    \label{fig:pariplot}
    \vspace{-3mm}
\end{figure}

\paragraph{Assumption validation: Concept transfer}
To assess the transferability of latent concepts,  we trained models on base unimodal data, novel multimodal mixed data, RGB data, and sketch data, and then show their pairwise distributions between latent $\mathbf{\hat{z}}$ and visual representations $\mathbf{x}$ in \Cref{fig:pariplot}. Significant overlap between base data and novel data distributions in different settings supports our assumption.

\section{Conclusions}

This paper introduces the cross-modal few-shot learning task, which addresses real-world scenarios involving multiple visual modalities with only a few labeled examples. 
Unlike classical few-shot learning, our task presents additional challenges due to the inherent variability in visual characteristics, structural properties, and domain gaps between modalities.
To overcome these challenges, we propose the generative transfer learning framework, which enables efficient knowledge transfer from abundant unimodal data to data-scarce multi-modal scenarios by estimating shared latent concepts from unimodal data and generalizing them to unseen modalities.
Extensive experiments on seven unique multi-modal datasets, demonstrated the superiority of the proposed framework by effectively disentangling modality-independent representations from in-modality disturbances. The results also suggest that the model can estimate latent concepts from vast unimodal data and generalize these concepts to unseen modalities using only a few available samples, much like human cognitive processes.

\bibliography{ICCV_Arxiv_Version}

\begin{thebibliography}{78}
\providecommand{\natexlab}[1]{#1}
\providecommand{\url}[1]{\texttt{#1}}
\expandafter\ifx\csname urlstyle\endcsname\relax
  \providecommand{\doi}[1]{doi: #1}\else
  \providecommand{\doi}{doi: \begingroup \urlstyle{rm}\Url}\fi

\bibitem[Alayrac et~al.(2022)Alayrac, Donahue, Luc, Miech, Barr, Hasson, Lenc, Mensch, Millican, Reynolds, et~al.]{alayrac2022flamingo}
Jean-Baptiste Alayrac, Jeff Donahue, Pauline Luc, Antoine Miech, Iain Barr, Yana Hasson, Karel Lenc, Arthur Mensch, Katherine Millican, Malcolm Reynolds, et~al.
\newblock Flamingo: a visual language model for few-shot learning.
\newblock \emph{Advances in neural information processing systems}, 35:\penalty0 23716--23736, 2022.

\bibitem[Bandyopadhyay et~al.(2024)Bandyopadhyay, Chowdhury, Sain, Koley, Xiang, Bhunia, and Song]{bandyopadhyay2024generalised}
Hmrishav Bandyopadhyay, Pinaki~Nath Chowdhury, Aneeshan Sain, Subhadeep Koley, Tao Xiang, Ayan~Kumar Bhunia, and Yi-Zhe Song.
\newblock Do generalised classifiers really work on human drawn sketches?
\newblock \emph{arXiv preprint arXiv:2407.03893}, 2024.

\bibitem[Bhunia et~al.(2022)Bhunia, Gajjala, Koley, Kundu, Sain, Xiang, and Song]{bhunia2022doodle}
Ayan~Kumar Bhunia, Viswanatha~Reddy Gajjala, Subhadeep Koley, Rohit Kundu, Aneeshan Sain, Tao Xiang, and Yi-Zhe Song.
\newblock Doodle it yourself: Class incremental learning by drawing a few sketches.
\newblock In \emph{Proceedings of the IEEE/CVF conference on computer vision and pattern recognition}, pages 2293--2302, 2022.

\bibitem[Cao et~al.(2021)Cao, Brbi\'c, and Leskovec]{cao2021concept}
Kaidi Cao, Maria Brbi\'c, and Jure Leskovec.
\newblock Concept learners for few-shot learning.
\newblock In \emph{International Conference on Learning Representations}, 2021.

\bibitem[Chen et~al.(2019)Chen, Liu, Kira, Wang, and Huang]{chen2019closer}
Wei{-}Yu Chen, Yen{-}Cheng Liu, Zsolt Kira, Yu{-}Chiang~Frank Wang, and Jia{-}Bin Huang.
\newblock A closer look at few-shot classification.
\newblock In \emph{International Conference on Learning Representations}. OpenReview.net, 2019.

\bibitem[Conti et~al.(2023)Conti, Fini, Mancini, Rota, Wang, and Ricci]{conti2023vocabulary}
Alessandro Conti, Enrico Fini, Massimiliano Mancini, Paolo Rota, Yiming Wang, and Elisa Ricci.
\newblock Vocabulary-free image classification.
\newblock \emph{Advances in Neural Information Processing Systems}, 36:\penalty0 30662--30680, 2023.

\bibitem[Deecke et~al.(2022)Deecke, Hospedales, and Bilen]{deecke2022visual}
Lucas Deecke, Timothy Hospedales, and Hakan Bilen.
\newblock Visual representation learning over latent domains.
\newblock In \emph{International Conference on Learning Representations}, 2022.

\bibitem[Deng et~al.(2009)Deng, Dong, Socher, Li, Li, and Fei-Fei]{deng2009imagenet}
Jia Deng, Wei Dong, Richard Socher, Li-Jia Li, Kai Li, and Li Fei-Fei.
\newblock Imagenet: A large-scale hierarchical image database.
\newblock In \emph{2009 IEEE conference on computer vision and pattern recognition}, pages 248--255. Ieee, 2009.

\bibitem[Dosovitskiy et~al.(2021)Dosovitskiy, Beyer, Kolesnikov, Weissenborn, Zhai, Unterthiner, Dehghani, Minderer, Heigold, Gelly, Uszkoreit, and Houlsby]{vit2020image}
Alexey Dosovitskiy, Lucas Beyer, Alexander Kolesnikov, Dirk Weissenborn, Xiaohua Zhai, Thomas Unterthiner, Mostafa Dehghani, Matthias Minderer, Georg Heigold, Sylvain Gelly, Jakob Uszkoreit, and Neil Houlsby.
\newblock An image is worth 16x16 words: Transformers for image recognition at scale.
\newblock In \emph{{ICLR}}. OpenReview.net, 2021.

\bibitem[Duan et~al.(2024)Duan, Xia, Mingze, Tang, Zhu, and Zhao]{duan2024cross}
Haoyi Duan, Yan Xia, Zhou Mingze, Li Tang, Jieming Zhu, and Zhou Zhao.
\newblock Cross-modal prompts: Adapting large pre-trained models for audio-visual downstream tasks.
\newblock \emph{Advances in Neural Information Processing Systems}, 36, 2024.

\bibitem[Eitz et~al.(2012)Eitz, Hays, and Alexa]{sketch12}
Mathias Eitz, James Hays, and Marc Alexa.
\newblock How do humans sketch objects?
\newblock \emph{ACM Trans. Graph.}, 31\penalty0 (4), 2012.

\bibitem[Fifty et~al.(2024)Fifty, Duan, Junkins, Amid, Leskovec, Re, and Thrun]{fifty2024contextaware}
Christopher Fifty, Dennis Duan, Ronald~Guenther Junkins, Ehsan Amid, Jure Leskovec, Christopher Re, and Sebastian Thrun.
\newblock Context-aware meta-learning.
\newblock In \emph{International Conference on Learning Representations}, 2024.

\bibitem[Fu et~al.(2023)Fu, Xie, Fu, and Jiang]{fu2023styleadv}
Yuqian Fu, Yu Xie, Yanwei Fu, and Yu-Gang Jiang.
\newblock Styleadv: Meta style adversarial training for cross-domain few-shot learning.
\newblock In \emph{Proceedings of the IEEE/CVF conference on computer vision and pattern recognition}, pages 24575--24584, 2023.

\bibitem[Guo et~al.(2020)Guo, Codella, Karlinsky, Codella, Smith, Saenko, Rosing, and Feris]{guo2020broader}
Yunhui Guo, Noel~C Codella, Leonid Karlinsky, James~V Codella, John~R Smith, Kate Saenko, Tajana Rosing, and Rogerio Feris.
\newblock A broader study of cross-domain few-shot learning.
\newblock In \emph{Computer vision--ECCV 2020: 16th European conference, glasgow, UK, August 23--28, 2020, proceedings, part XXVII 16}, pages 124--141. Springer, 2020.

\bibitem[Han et~al.(2024)Han, Yoon, Arik, and Pfister]{han2024large}
Sungwon Han, Jinsung Yoon, Sercan~O Arik, and Tomas Pfister.
\newblock Large language models can automatically engineer features for few-shot tabular learning.
\newblock \emph{arXiv preprint arXiv:2404.09491}, 2024.

\bibitem[Han et~al.(2023)Han, Han, Liu, Wang, Pan, Pu, Deng, Feng, Song, and Huang]{han2023dynamic}
Yizeng Han, Dongchen Han, Zeyu Liu, Yulin Wang, Xuran Pan, Yifan Pu, Chao Deng, Junlan Feng, Shiji Song, and Gao Huang.
\newblock Dynamic perceiver for efficient visual recognition.
\newblock In \emph{Proceedings of the IEEE/CVF International Conference on Computer Vision}, pages 5992--6002, 2023.

\bibitem[He et~al.(2021)He, Luo, Wang, Wang, Li, and Jiang]{he2021transreid}
Shuting He, Hao Luo, Pichao Wang, Fan Wang, Hao Li, and Wei Jiang.
\newblock Transreid: Transformer-based object re-identification.
\newblock In \emph{Proceedings of the IEEE/CVF international conference on computer vision}, pages 15013--15022, 2021.

\bibitem[Hu et~al.(2022)Hu, Wang, Sun, Wang, and Xue]{hu2022m}
Pingyi Hu, Zihan Wang, Ruoxi Sun, Hu Wang, and Minhui Xue.
\newblock M4 i: Multi-modal models membership inference.
\newblock \emph{Advances in Neural Information Processing Systems}, 35:\penalty0 1867--1882, 2022.

\bibitem[Jiang et~al.(2023{\natexlab{a}})Jiang, Xu, Xin, Liang, Peng, Zhang, and Zhu]{jiang2023mewl}
Guangyuan Jiang, Manjie Xu, Shiji Xin, Wei Liang, Yujia Peng, Chi Zhang, and Yixin Zhu.
\newblock Mewl: Few-shot multimodal word learning with referential uncertainty.
\newblock In \emph{International Conference on Machine Learning}, pages 15144--15169. PMLR, 2023{\natexlab{a}}.

\bibitem[Jiang et~al.(2023{\natexlab{b}})Jiang, Chen, Zhao, Chen, Ping, Tran, Xu, Zeng, and Chilimbi]{jiang2023understanding}
Qian Jiang, Changyou Chen, Han Zhao, Liqun Chen, Qing Ping, Son~Dinh Tran, Yi Xu, Belinda Zeng, and Trishul Chilimbi.
\newblock Understanding and constructing latent modality structures in multi-modal representation learning.
\newblock In \emph{Proceedings of the IEEE/CVF Conference on Computer Vision and Pattern Recognition}, pages 7661--7671, 2023{\natexlab{b}}.

\bibitem[Ju et~al.(2022)Ju, Han, Zheng, Zhang, and Xie]{ju2022prompting}
Chen Ju, Tengda Han, Kunhao Zheng, Ya Zhang, and Weidi Xie.
\newblock Prompting visual-language models for efficient video understanding.
\newblock In \emph{European Conference on Computer Vision}, pages 105--124. Springer, 2022.

\bibitem[Karras et~al.(2018)Karras, Aila, Laine, and Lehtinen]{Tero18CelebHQ}
Tero Karras, Timo Aila, Samuli Laine, and Jaakko Lehtinen.
\newblock Progressive growing of gans for improved quality, stability, and variation.
\newblock In \emph{International Conference on Learning Representations}, 2018.

\bibitem[Ke et~al.(2024)Ke, Cao, Ling, and Zhou]{ke2024revisiting}
Tianjun Ke, Haoqun Cao, Zenan Ling, and Feng Zhou.
\newblock Revisiting logistic-softmax likelihood in bayesian meta-learning for few-shot classification.
\newblock \emph{Advances in Neural Information Processing Systems}, 36, 2024.

\bibitem[Khosla et~al.(2011)Khosla, Jayadevaprakash, Yao, and Li]{khosla2011novel}
Aditya Khosla, Nityananda Jayadevaprakash, Bangpeng Yao, and Fei-Fei Li.
\newblock Novel dataset for fine-grained image categorization: Stanford dogs.
\newblock In \emph{Proc. CVPR workshop on fine-grained visual categorization (FGVC)}, 2011.

\bibitem[Kingma and Ba(2015)]{KingmaB14}
Diederik~P. Kingma and Jimmy Ba.
\newblock Adam: {A} method for stochastic optimization.
\newblock In \emph{Proc. Int. Conf. Learn. Represent.}, pages 1--15, 2015.

\bibitem[Kong et~al.(2024)Kong, Goel, Badlani, Ping, Valle, and Catanzaro]{kong2024audio}
Zhifeng Kong, Arushi Goel, Rohan Badlani, Wei Ping, Rafael Valle, and Bryan Catanzaro.
\newblock Audio flamingo: A novel audio language model with few-shot learning and dialogue abilities.
\newblock \emph{arXiv preprint arXiv:2402.01831}, 2024.

\bibitem[Lai et~al.(2011)Lai, Bo, Ren, and Fox]{lai2011large}
Kevin Lai, Liefeng Bo, Xiaofeng Ren, and Dieter Fox.
\newblock A large-scale hierarchical multi-view rgb-d object dataset.
\newblock In \emph{2011 IEEE international conference on robotics and automation}, pages 1817--1824. IEEE, 2011.

\bibitem[Lake and Piantadosi(2020)]{lake2020people}
Brenden~M Lake and Steven~T Piantadosi.
\newblock People infer recursive visual concepts from just a few examples.
\newblock \emph{Computational Brain \& Behavior}, 3\penalty0 (1):\penalty0 54--65, 2020.

\bibitem[Li et~al.(2020)Li, Zhang, Li, and Fu]{li2020adversarial}
Kai Li, Yulun Zhang, Kunpeng Li, and Yun Fu.
\newblock Adversarial feature hallucination networks for few-shot learning.
\newblock In \emph{Proceedings of the IEEE/CVF conference on computer vision and pattern recognition}, pages 13470--13479, 2020.

\bibitem[Li et~al.(2023{\natexlab{a}})Li, Wang, Zhang, Gao, Song, Liu, Li, and Qiao]{Li2023uniformer}
Kunchang Li, Yali Wang, Junhao Zhang, Peng Gao, Guanglu Song, Yu Liu, Hongsheng Li, and Yu Qiao.
\newblock Uniformer: Unifying convolution and self-attention for visual recognition.
\newblock \emph{IEEE Transactions on Pattern Analysis and Machine Intelligence}, 45\penalty0 (10):\penalty0 12581--12600, 2023{\natexlab{a}}.

\bibitem[Li et~al.(2024{\natexlab{a}})Li, Xiao, Chen, Shao, Zhuang, and Chen]{li2024zero}
Lin Li, Jun Xiao, Guikun Chen, Jian Shao, Yueting Zhuang, and Long Chen.
\newblock Zero-shot visual relation detection via composite visual cues from large language models.
\newblock \emph{Advances in Neural Information Processing Systems}, 36, 2024{\natexlab{a}}.

\bibitem[Li et~al.(2023{\natexlab{b}})Li, Sun, and Li]{li2023clip}
Siyuan Li, Li Sun, and Qingli Li.
\newblock Clip-reid: exploiting vision-language model for image re-identification without concrete text labels.
\newblock In \emph{Proceedings of the AAAI Conference on Artificial Intelligence}, pages 1405--1413, 2023{\natexlab{b}}.

\bibitem[Li et~al.(2022)Li, Liu, and Bilen]{Li_2022_CVPR}
Wei-Hong Li, Xialei Liu, and Hakan Bilen.
\newblock Cross-domain few-shot learning with task-specific adapters.
\newblock In \emph{Proceedings of the IEEE/CVF Conference on Computer Vision and Pattern Recognition}, pages 7161--7170, 2022.

\bibitem[Li et~al.(2024{\natexlab{b}})Li, Chen, Abramowitz, Anzellotti, and Wei]{lilearning}
Yuke Li, Guangyi Chen, Ben Abramowitz, Stefano Anzellotti, and Donglai Wei.
\newblock Learning causal domain-invariant temporal dynamics for few-shot action recognition.
\newblock In \emph{Forty-first International Conference on Machine Learning}, 2024{\natexlab{b}}.

\bibitem[Lin et~al.(2023{\natexlab{a}})Lin, Wang, Wang, Zheng, and Satoh]{lin2023beyond}
Kejun Lin, Z. Wang, Zheng Wang, Yinqiang Zheng, and Shin’ichi Satoh.
\newblock Beyond domain gap: Exploiting subjectivity in sketch-based person retrieval.
\newblock In \emph{Proceedings of the ACM International Conference on Multimedia}, pages 2078--2089, 2023{\natexlab{a}}.

\bibitem[Lin et~al.(2023{\natexlab{b}})Lin, Sung, Lei, Bansal, and Bertasius]{Lin_2023_CVPR}
Yan-Bo Lin, Yi-Lin Sung, Jie Lei, Mohit Bansal, and Gedas Bertasius.
\newblock Vision transformers are parameter-efficient audio-visual learners.
\newblock In \emph{Proceedings of the IEEE/CVF Conference on Computer Vision and Pattern Recognition (CVPR)}, pages 2299--2309, 2023{\natexlab{b}}.

\bibitem[Lin et~al.(2023{\natexlab{c}})Lin, Yu, Kuang, Pathak, and Ramanan]{lin2023multimodality}
Zhiqiu Lin, Samuel Yu, Zhiyi Kuang, Deepak Pathak, and Deva Ramanan.
\newblock Multimodality helps unimodality: Cross-modal few-shot learning with multimodal models.
\newblock In \emph{Proceedings of the IEEE/CVF Conference on Computer Vision and Pattern Recognition}, pages 19325--19337, 2023{\natexlab{c}}.

\bibitem[Liu et~al.(2017)Liu, Hu, and Ma]{liu2017online}
Hong Liu, Liang Hu, and Liqian Ma.
\newblock Online rgb-d person re-identification based on metric model update.
\newblock \emph{CAAI Transactions on Intelligence Technology}, 2\penalty0 (1):\penalty0 48--55, 2017.

\bibitem[Liu et~al.(2022)Liu, Yuan, Dai, Shen, Zhu, Chen, and He]{liu2022few}
Ruixue Liu, Shaozu Yuan, Aijun Dai, Lei Shen, Tiangang Zhu, Meng Chen, and Xiaodong He.
\newblock Few-shot table understanding: A benchmark dataset and pre-training baseline.
\newblock In \emph{Proceedings of the 29th International Conference on Computational Linguistics}, pages 3741--3752, 2022.

\bibitem[Luo et~al.(2023)Luo, Wu, Zhang, Gao, Xu, and Song]{luo2023closer}
Xu Luo, Hao Wu, Ji Zhang, Lianli Gao, Jing Xu, and Jingkuan Song.
\newblock A closer look at few-shot classification again.
\newblock In \emph{International Conference on Machine Learning}, pages 23103--23123. PMLR, 2023.

\bibitem[Ma et~al.(2024)Ma, Chen, Zhao, Zhang, Luo, and Xu]{ma2024cross}
Zhen-Xiang Ma, Zhen-Duo Chen, Li-Jun Zhao, Zi-Chao Zhang, Xin Luo, and Xin-Shun Xu.
\newblock Cross-layer and cross-sample feature optimization network for few-shot fine-grained image classification.
\newblock In \emph{Proceedings of the AAAI Conference on Artificial Intelligence}, pages 4136--4144, 2024.

\bibitem[Majumder et~al.(2022)Majumder, Chen, Al-Halah, and Grauman]{majumder2022few}
Sagnik Majumder, Changan Chen, Ziad Al-Halah, and Kristen Grauman.
\newblock Few-shot audio-visual learning of environment acoustics.
\newblock \emph{Advances in Neural Information Processing Systems}, 35:\penalty0 2522--2536, 2022.

\bibitem[Meng et~al.(2023)Meng, Michalski, Huang, Zhang, Abdelzaher, and Han]{meng2023tuning}
Yu Meng, Martin Michalski, Jiaxin Huang, Yu Zhang, Tarek Abdelzaher, and Jiawei Han.
\newblock Tuning language models as training data generators for augmentation-enhanced few-shot learning.
\newblock In \emph{International Conference on Machine Learning}, pages 24457--24477. PMLR, 2023.

\bibitem[Meshry et~al.(2021)Meshry, Suri, Davis, and Shrivastava]{meshry2021learned}
Moustafa Meshry, Saksham Suri, Larry~S Davis, and Abhinav Shrivastava.
\newblock Learned spatial representations for few-shot talking-head synthesis.
\newblock In \emph{Proceedings of the IEEE/CVF international conference on computer vision}, pages 13829--13838, 2021.

\bibitem[Mirza et~al.(2024)Mirza, Karlinsky, Lin, Possegger, Kozinski, Feris, and Bischof]{mirza2024lafter}
Muhammad~Jehanzeb Mirza, Leonid Karlinsky, Wei Lin, Horst Possegger, Mateusz Kozinski, Rogerio Feris, and Horst Bischof.
\newblock Lafter: Label-free tuning of zero-shot classifier using language and unlabeled image collections.
\newblock \emph{Advances in Neural Information Processing Systems}, 36, 2024.

\bibitem[Mok et~al.(2024)Mok, Li, Bai, Zhang, Liu, Zhou, Yan, Jin, Shi, Yin, et~al.]{mok2024modality}
Tony~CW Mok, Zi Li, Yunhao Bai, Jianpeng Zhang, Wei Liu, Yan-Jie Zhou, Ke Yan, Dakai Jin, Yu Shi, Xiaoli Yin, et~al.
\newblock Modality-agnostic structural image representation learning for deformable multi-modality medical image registration.
\newblock In \emph{Proceedings of the IEEE/CVF Conference on Computer Vision and Pattern Recognition}, pages 11215--11225, 2024.

\bibitem[Mukherjee et~al.(2024)Mukherjee, Huey, Lu, Vinker, Aguina-Kang, Shamir, and Fan]{mukherjee2024seva}
Kushin Mukherjee, Holly Huey, Xuanchen Lu, Yael Vinker, Rio Aguina-Kang, Ariel Shamir, and Judith Fan.
\newblock Seva: Leveraging sketches to evaluate alignment between human and machine visual abstraction.
\newblock \emph{Advances in Neural Information Processing Systems}, 36, 2024.

\bibitem[Oreshkin et~al.(2018)Oreshkin, Rodr{\'\i}guez~L{\'o}pez, and Lacoste]{oreshkin2018tadam}
Boris Oreshkin, Pau Rodr{\'\i}guez~L{\'o}pez, and Alexandre Lacoste.
\newblock Tadam: Task dependent adaptive metric for improved few-shot learning.
\newblock \emph{Advances in neural information processing systems}, 31, 2018.

\bibitem[Qing et~al.(2023)Qing, Zhang, Huang, Zhang, Gao, Zhao, and Sang]{qing2023disentangling}
Zhiwu Qing, Shiwei Zhang, Ziyuan Huang, Yingya Zhang, Changxin Gao, Deli Zhao, and Nong Sang.
\newblock Disentangling spatial and temporal learning for efficient image-to-video transfer learning.
\newblock In \emph{Proceedings of the IEEE/CVF International Conference on Computer Vision}, pages 13934--13944, 2023.

\bibitem[Radford et~al.(2021)Radford, Kim, Hallacy, Ramesh, Goh, Agarwal, Sastry, Askell, Mishkin, Clark, et~al.]{radford2021clip}
Alec Radford, Jong~Wook Kim, Chris Hallacy, Aditya Ramesh, Gabriel Goh, Sandhini Agarwal, Girish Sastry, Amanda Askell, Pamela Mishkin, Jack Clark, et~al.
\newblock Learning transferable visual models from natural language supervision.
\newblock In \emph{International conference on machine learning}, pages 8748--8763. PMLR, 2021.

\bibitem[Sangkloy et~al.(2016)Sangkloy, Burnell, Ham, and Hays]{sangkloy2016sketchy}
Patsorn Sangkloy, Nathan Burnell, Cusuh Ham, and James Hays.
\newblock The sketchy database: learning to retrieve badly drawn bunnies.
\newblock \emph{ACM Transactions on Graphics (TOG)}, 35\penalty0 (4):\penalty0 1--12, 2016.

\bibitem[Shao et~al.(2024)Shao, Bai, Wang, Liu, and Liu]{shao2024collaborative}
Shuai Shao, Yu Bai, Yan Wang, Baodi Liu, and Bin Liu.
\newblock Collaborative consortium of foundation models for open-world few-shot learning.
\newblock In \emph{Proceedings of the AAAI Conference on Artificial Intelligence}, pages 4740--4747, 2024.

\bibitem[Sheng et~al.(2024)Sheng, Kuang, Bai, Hou, Guo, Xu, Pietik{\"a}inen, and Liu]{sheng2024deep}
Changchong Sheng, Gangyao Kuang, Liang Bai, Chenping Hou, Yulan Guo, Xin Xu, Matti Pietik{\"a}inen, and Li Liu.
\newblock Deep learning for visual speech analysis: A survey.
\newblock \emph{IEEE Transactions on Pattern Analysis and Machine Intelligence}, 2024.

\bibitem[Song et~al.(2023)Song, Xue, Wang, Sun, Ge, Shan, et~al.]{song2023meta}
Lin Song, Ruoyi Xue, Hang Wang, Hongbin Sun, Yixiao Ge, Ying Shan, et~al.
\newblock Meta-adapter: An online few-shot learner for vision-language model.
\newblock \emph{Advances in Neural Information Processing Systems}, 36:\penalty0 55361--55374, 2023.

\bibitem[Sun et~al.(2019)Sun, Liu, Chua, and Schiele]{sun2019meta}
Qianru Sun, Yaoyao Liu, Tat-Seng Chua, and Bernt Schiele.
\newblock Meta-transfer learning for few-shot learning.
\newblock In \emph{Proceedings of the IEEE/CVF conference on computer vision and pattern recognition}, pages 403--412, 2019.

\bibitem[Tharwat and Schenck(2023)]{tharwat2023survey}
Alaa Tharwat and Wolfram Schenck.
\newblock A survey on active learning: State-of-the-art, practical challenges and research directions.
\newblock \emph{Mathematics}, 11\penalty0 (4):\penalty0 820, 2023.

\bibitem[Tian et~al.(2020)Tian, Wang, Krishnan, Tenenbaum, and Isola]{tian2020rethinking}
Yonglong Tian, Yue Wang, Dilip Krishnan, Joshua~B Tenenbaum, and Phillip Isola.
\newblock Rethinking few-shot image classification: a good embedding is all you need?
\newblock In \emph{Computer Vision--ECCV 2020: 16th European Conference, Glasgow, UK, August 23--28, 2020, Proceedings, Part XIV 16}, pages 266--282. Springer, 2020.

\bibitem[Tsimpoukelli et~al.(2021)Tsimpoukelli, Menick, Cabi, Eslami, Vinyals, and Hill]{tsimpoukelli2021multimodal}
Maria Tsimpoukelli, Jacob~L Menick, Serkan Cabi, SM Eslami, Oriol Vinyals, and Felix Hill.
\newblock Multimodal few-shot learning with frozen language models.
\newblock \emph{Advances in Neural Information Processing Systems}, 34:\penalty0 200--212, 2021.

\bibitem[Vinker et~al.(2022)Vinker, Pajouheshgar, Bo, Bachmann, Bermano, Cohen-Or, Zamir, and Shamir]{vinker2022clipasso}
Yael Vinker, Ehsan Pajouheshgar, Jessica~Y Bo, Roman~Christian Bachmann, Amit~Haim Bermano, Daniel Cohen-Or, Amir Zamir, and Ariel Shamir.
\newblock Clipasso: Semantically-aware object sketching.
\newblock \emph{ACM Transactions on Graphics (TOG)}, 41\penalty0 (4):\penalty0 1--11, 2022.

\bibitem[Vinker et~al.(2023)Vinker, Alaluf, Cohen-Or, and Shamir]{vinker2023clipascene}
Yael Vinker, Yuval Alaluf, Daniel Cohen-Or, and Ariel Shamir.
\newblock Clipascene: Scene sketching with different types and levels of abstraction.
\newblock In \emph{Proceedings of the IEEE/CVF International Conference on Computer Vision}, pages 4146--4156, 2023.

\bibitem[Wah et~al.(2011)Wah, Branson, Welinder, Perona, and Belongie]{wah2011caltech}
Catherine Wah, Steve Branson, Peter Welinder, Pietro Perona, and Serge Belongie.
\newblock The caltech-ucsd birds-200-2011 dataset.
\newblock In \emph{California Institute of Technology}, 2011.

\bibitem[Wang et~al.(2020)Wang, Song, Wei, and Zhang]{wang2020adversarial}
Yimu Wang, Renjie Song, Xiu-Shen Wei, and Lijun Zhang.
\newblock An adversarial domain adaptation network for cross-domain fine-grained recognition.
\newblock In \emph{Proceedings of the IEEE/CVF Winter Conference on Applications of Computer Vision}, pages 1228--1236, 2020.

\bibitem[Wei et~al.(2018)Wei, Zhang, Gao, and Tian]{wei2018person}
Longhui Wei, Shiliang Zhang, Wen Gao, and Qi Tian.
\newblock Person transfer gan to bridge domain gap for person re-identification.
\newblock In \emph{Proceedings of the IEEE conference on computer vision and pattern recognition}, pages 79--88, 2018.

\bibitem[Wei et~al.(2023)Wei, Cao, Zhang, Peng, Yao, Xie, Hu, and Guo]{Wei_2023_CVPR}
Yixuan Wei, Yue Cao, Zheng Zhang, Houwen Peng, Zhuliang Yao, Zhenda Xie, Han Hu, and Baining Guo.
\newblock iclip: Bridging image classification and contrastive language-image pre-training for visual recognition.
\newblock In \emph{Proceedings of the IEEE/CVF Conference on Computer Vision and Pattern Recognition (CVPR)}, pages 2776--2786, 2023.

\bibitem[Wu et~al.(2017)Wu, Zheng, Yu, Gong, and Lai]{wu2017rgb}
Ancong Wu, Wei-Shi Zheng, Hong-Xing Yu, Shaogang Gong, and Jianhuang Lai.
\newblock Rgb-infrared cross-modality person re-identification.
\newblock In \emph{Proceedings of the IEEE international conference on computer vision}, pages 5380--5389, 2017.

\bibitem[Wu et~al.(2024)Wu, Zheng, Ren, Vasluianu, Ma, Paudel, Van~Gool, and Timofte]{wu2024single}
Zongwei Wu, Jilai Zheng, Xiangxuan Ren, Florin-Alexandru Vasluianu, Chao Ma, Danda~Pani Paudel, Luc Van~Gool, and Radu Timofte.
\newblock Single-model and any-modality for video object tracking.
\newblock In \emph{Proceedings of the IEEE/CVF Conference on Computer Vision and Pattern Recognition}, pages 19156--19166, 2024.

\bibitem[Xing et~al.(2019)Xing, Rostamzadeh, Oreshkin, and O~Pinheiro]{xing2019adaptive}
Chen Xing, Negar Rostamzadeh, Boris Oreshkin, and Pedro~O O~Pinheiro.
\newblock Adaptive cross-modal few-shot learning.
\newblock \emph{Advances in neural information processing systems}, 32, 2019.

\bibitem[Xu et~al.(2023)Xu, Zhi, Sun, Patel, and Liu]{xu2023deep}
Huali Xu, Shuaifeng Zhi, Shuzhou Sun, Vishal~M Patel, and Li Liu.
\newblock Deep learning for cross-domain few-shot visual recognition: A survey.
\newblock \emph{arXiv preprint arXiv:2303.08557}, 2023.

\bibitem[Yan et~al.(2023{\natexlab{a}})Yan, Wang, Zhong, Dong, He, Lu, Wang, Shang, and McAuley]{yan2023learning}
An Yan, Yu Wang, Yiwu Zhong, Chengyu Dong, Zexue He, Yujie Lu, William~Yang Wang, Jingbo Shang, and Julian McAuley.
\newblock Learning concise and descriptive attributes for visual recognition.
\newblock In \emph{Proceedings of the IEEE/CVF International Conference on Computer Vision}, pages 3090--3100, 2023{\natexlab{a}}.

\bibitem[Yan et~al.(2023{\natexlab{b}})Yan, Zhang, Chen, Tang, Zhu, Sun, Van~Gool, and Zhang]{yan2023smae}
Qingsen Yan, Song Zhang, Weiye Chen, Hao Tang, Yu Zhu, Jinqiu Sun, Luc Van~Gool, and Yanning Zhang.
\newblock Smae: Few-shot learning for hdr deghosting with saturation-aware masked autoencoders.
\newblock In \emph{Proceedings of the IEEE/CVF Conference on Computer Vision and Pattern Recognition}, pages 5775--5784, 2023{\natexlab{b}}.

\bibitem[Ye et~al.(2024)Ye, Liu, Cai, Zhou, and Zhan]{ye2024closerlookdeeplearning}
Han-Jia Ye, Si-Yang Liu, Hao-Run Cai, Qi-Le Zhou, and De-Chuan Zhan.
\newblock A closer look at deep learning on tabular data, 2024.

\bibitem[Ye et~al.(2021)Ye, Shen, Lin, Xiang, Shao, and Hoi]{ye2021deep}
Mang Ye, Jianbing Shen, Gaojie Lin, Tao Xiang, Ling Shao, and Steven~CH Hoi.
\newblock Deep learning for person re-identification: A survey and outlook.
\newblock \emph{IEEE transactions on pattern analysis and machine intelligence}, 44\penalty0 (6):\penalty0 2872--2893, 2021.

\bibitem[Yun et~al.(2024)Yun, Seo, Seo, Yoon, Kim, Ji, Ashtari, and Noh]{yun2024stylized}
Kwan Yun, Kwanggyoon Seo, Chang~Wook Seo, Soyeon Yoon, Seongcheol Kim, Soohyun Ji, Amirsaman Ashtari, and Junyong Noh.
\newblock Stylized face sketch extraction via generative prior with limited data.
\newblock In \emph{Computer Graphics Forum}, page e15045. Wiley Online Library, 2024.

\bibitem[Zhang et~al.(2016)Zhang, Liu, Zhang, Ren, Wang, and Cao]{zhang2016sketchnet}
Hua Zhang, Si Liu, Changqing Zhang, Wenqi Ren, Rui Wang, and Xiaochun Cao.
\newblock Sketchnet: Sketch classification with web images.
\newblock In \emph{Proceedings of the IEEE conference on computer vision and pattern recognition}, pages 1105--1113, 2016.

\bibitem[Zhang et~al.(2024)Zhang, Li, Wu, and Kuang]{zhang24metacoco}
Min Zhang, Haoxuan Li, Fei Wu, and Kun Kuang.
\newblock Metacoco: {A} new few-shot classification benchmark with spurious correlation.
\newblock In \emph{{ICLR}}. OpenReview.net, 2024.

\bibitem[Zheng et~al.(2015)Zheng, Shen, Tian, Wang, Wang, and Tian]{zheng2015scalable}
Liang Zheng, Liyue Shen, Lu Tian, Shengjin Wang, Jingdong Wang, and Qi Tian.
\newblock Scalable person re-identification: A benchmark.
\newblock In \emph{Proceedings of the IEEE international conference on computer vision}, pages 1116--1124, 2015.

\bibitem[Zhou et~al.(2023)Zhou, Wang, Zhang, Wei, and Zhang]{zhou2023revisiting}
Fei Zhou, Peng Wang, Lei Zhang, Wei Wei, and Yanning Zhang.
\newblock Revisiting prototypical network for cross domain few-shot learning.
\newblock In \emph{Proceedings of the IEEE/CVF conference on computer vision and pattern recognition}, pages 20061--20070, 2023.

\bibitem[Zhu et~al.(2024)Zhu, Chen, Ji, Ye, and Liu]{zhu2024llafs}
Lanyun Zhu, Tianrun Chen, Deyi Ji, Jieping Ye, and Jun Liu.
\newblock Llafs: When large language models meet few-shot segmentation.
\newblock In \emph{Proceedings of the IEEE/CVF Conference on Computer Vision and Pattern Recognition}, pages 3065--3075, 2024.

\end{thebibliography}
\bibliographystyle{ieeenat_fullname}

\appendix

\section{Related Work}
\label{apendix:related}
\subsection{Relation to Unimodal Few-shot Learning} 
In general, unimodal learning tasks can be broadly categorized into three types based on the availability and quantity of labeled data:
(i) {\bf Supervised Learning}, where a large amount of labeled data is available for training, allowing models to learn features and perform accurate recognition~\citep{yan2023learning, han2023dynamic, Li2023uniformer}; 
(ii) {\bf Few-shot Learning}, where only a limited number of labeled samples are provided for each class, challenging the model to generalize effectively from minimal data~\citep{chen2019closer, luo2023closer}; 
and (iii) {\bf Zero-shot Learning}, where no labeled examples are available for certain classes~\citep{Wei_2023_CVPR, li2024zero, mirza2024lafter}. 

Specifically, Few-shot learning (FSL) encompasses a training phase where a model is trained on a relatively large dataset and an adaptation phase in which the trained model is adjusted to previously unseen tasks with limited labeled samples.
Most existing FSL tasks utilize unimodal datasets for training and testing, including popular benchmarks such as ImageNet~\citep{deng2009imagenet}, CIFAR~\citep{oreshkin2018tadam}, CUB-200-2021~\citep{wah2011caltech}, and Stanford Dogs~\citep{khosla2011novel}.
FSL typically involves three main approaches: (i) {\bf Meta learning}~\citep{sun2019meta, ma2024cross}, or learning to learn, which optimizes model parameters across diverse learning tasks to enable rapid adaptation to new challenges; (ii) {\bf Data-centric learning}~\citep{li2020adversarial, meng2023tuning, ma2024cross}, which focuses on metric learning to compare distances between samples or expand synthetic data facing with data-scarce scenarios; (iii) {\bf Transfer learning}~\citep{tian2020rethinking, luo2023closer, zhang24metacoco} where models pre-trained on large-scale datasets are fine-tuned on few-shot tasks to improve performance to improve performance by leveraging learned representations for more efficient adaptation.

The meta-learning methods~\citep{sun2019meta, cao2021concept, ma2024cross, fifty2024contextaware} are densely connected to the model design, which attempts to directly establish a mapping function between input and prediction. 
\citet{cao2021concept} separate targets as fine-grained parts to build concept learners. \citet{ma2024cross} incorporate semantic information and spatial position into a reconstruction framework to strengthen the feature alignment between seen and unseen samples.
By rapidly updating parameters on new tasks with a small number of samples, these methods facilitate the transfer of knowledge from previously learned tasks, making them highly effective in few-shot learning scenarios. 

The data-centric methods utilize synthetic data~\citep{meng2023tuning} or metric learning~\citep{ma2024cross} to adapt data-insufficient scenarios.
The former involves generation methods like Generative Adversarial Networks (GAN)~\citep{li2020adversarial} and auto-encoders~\citep{yan2023smae}. These approaches generate additional training data to enhance the model's performance, mitigating the scarcity of labeled examples and improving generalization to new tasks. 
The latter, metric learning, tries to build data connections with the thoughts of nearby neighbors, focusing on learning a distance metric that clusters similar examples together and separates dissimilar ones. These data-centric methods emphasize the properties of the data to enhance the model's ability to generalize from a few examples.

The fine-tuning-based methods~\citep{tian2020rethinking, luo2023closer, zhang24metacoco} involve using pre-trained models on large datasets and fine-tuning them on the target task with a small number of examples. 
This approach leverages the knowledge gained from the pre-training phase and transfers it to the specific requirements of the new task. 
However, with the increasing amount of raw data on the Internet, it's challenging for pre-trained models, including Vision-Language Pretraining Models (VLM)~\citep{zhu2024llafs}, to generalize to specific novel data, especially when the data is in different modalities. 

Recent works on cross-domain learning~\citep{wang2020adversarial, Li_2022_CVPR, xu2023deep} also focus exclusively on learning from unimodal data. However, their limitations become apparent in complex real-world applications that often require understanding multiple modalities simultaneously. 
In this work, we extend its applicability of few-shot learning to real-world scenarios where data come from diverse visual modalities.

\begin{figure*}[t]
    \centering
    \begin{subfigure}{0.4\linewidth}
        \centering
        \captionsetup{font=footnotesize, labelfont=footnotesize}
        \includegraphics[width=0.99\linewidth]{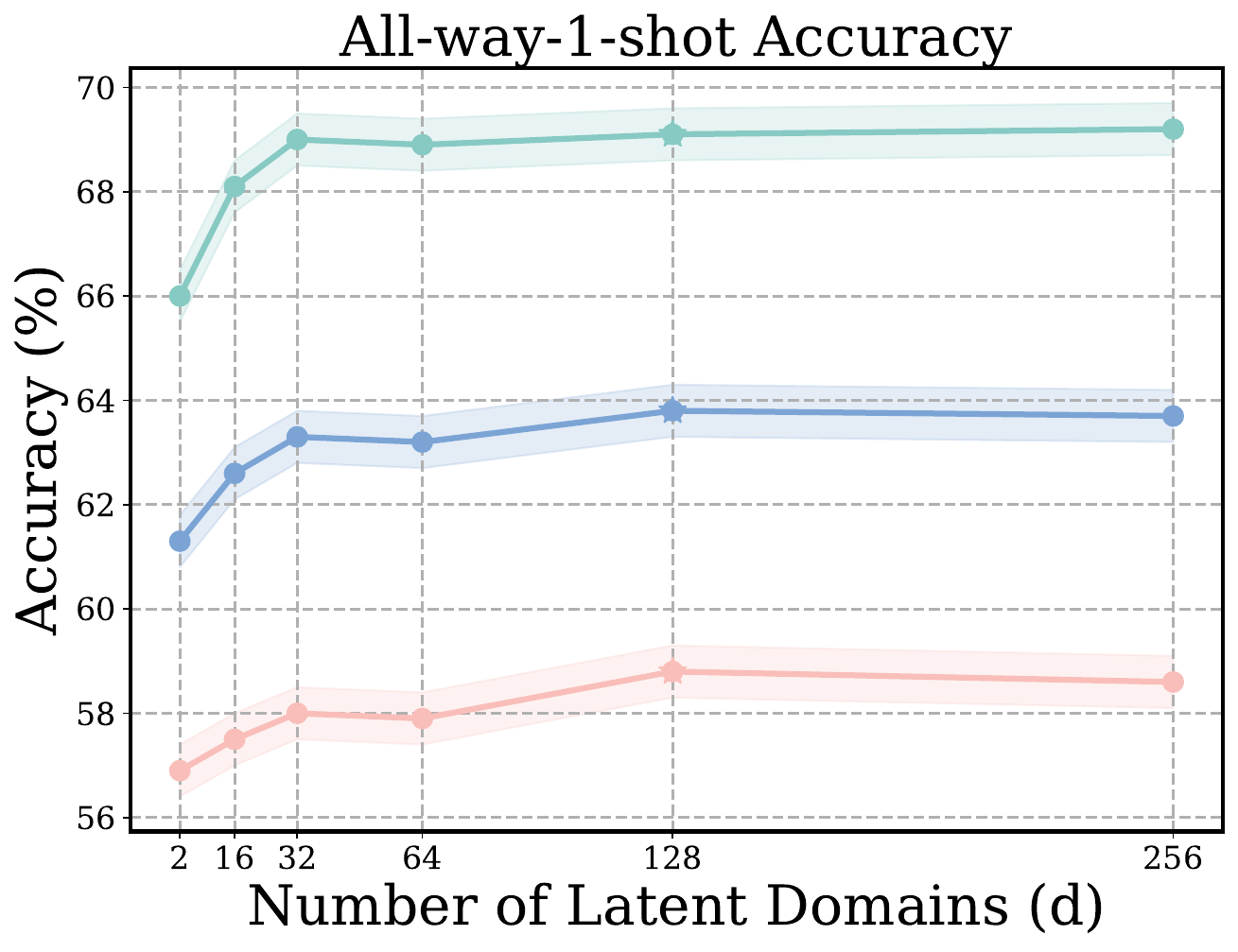}

    \end{subfigure}
    \begin{subfigure}{0.4\linewidth}

        \captionsetup{font=footnotesize, labelfont=footnotesize}
        \includegraphics[width=0.99\linewidth]{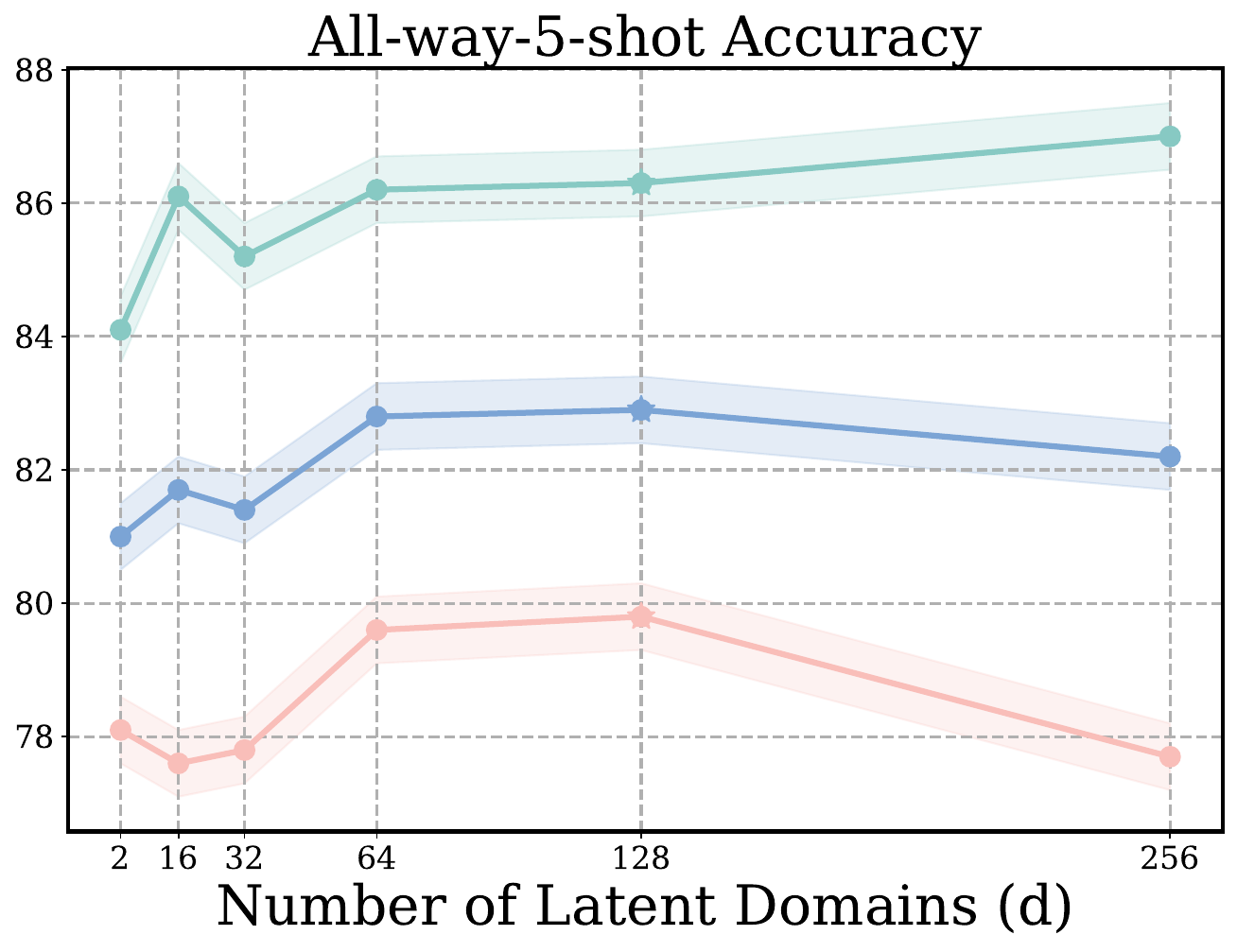}
    \end{subfigure}
    \caption{Hyperparameter analysis of the selection of the latent domain number $d$ on the \textsc{Sketchy} dataset. The \textcolor{blue}{Blue} line denotes the average accuracy of mixed modality data; the \textcolor{YellowGreen}{Green} line stands for the average accuracy of RGB data; the \textcolor{red}{RED} line represents the average accuracy of sketch data.}
    \label{fig:latentdomain}
\end{figure*}

\subsection{Relation to Multi-modal Few-shot Learning}
Existing multi-modal few-shot learning methods aim to leverage information from multiple modalities, such as combining visual data with textual~\citep{xing2019adaptive, tsimpoukelli2021multimodal, alayrac2022flamingo, lin2023multimodality, shao2024collaborative}, audio~\citep{meshry2021learned, majumder2022few, kong2024audio}, or tabular data~\cite{liu2022few, ye2024closerlookdeeplearning, han2024large}.
These approaches focus on enriching the feature space by integrating heterogeneous data sources, thereby improving the generalization capabilities of few-shot models in scenarios where additional modality information is available.
However, most of these methods concentrate on leveraging modalities that are inherently different from visual data, like text or audio, to provide semantic context or supplementary cues. They often assume access to richly annotated data in these auxiliary modalities, which may not always be feasible in practical applications. These methods enhance recognition by supplementing visual features with external information but do not address the challenges that arise when dealing with multiple visual modalities.

Few methods address the challenges of visual multi-modal scenarios where multiple visual modalities of the same object exist, such as RGB images, sketches, infrared images, or depth maps. Visual modalities often exhibit unique characteristics and structural properties, leading to significant domain gaps that complicate direct knowledge transfer between them. Some approaches~\citet{bhunia2022doodle} utilize additional visual modalities to enhance recognition performance on a primary modality; however, they typically treat the additional visual modality as auxiliary information to support unimodal recognition rather than fully integrating multiple visual modalities into a unified learning framework.

In contrast, our work specifically targets the problem of Cross-modal Few-Shot Learning (CFSL) within the domain of visual data. We focus on recognizing instances across different visual modalities when only a few labeled examples are available. Our approach acknowledges that in real-world scenarios, models must often adapt to new visual modalities with limited annotated data, without the luxury of abundant multi-modal annotations. Unlike existing methods that leverage auxiliary modalities to aid a primary modality, we aim to develop a model capable of understanding and generalizing across diverse visual modalities in a few-shot setting.

In summary, while existing multi-modal FSL methods aim to enhance performance by integrating different data types, our work addresses the unique challenges of cross-modal recognition within visual data. We emphasize the importance of transferring knowledge from abundant unimodal data to novel visual modalities in a few-shot context, without relying on auxiliary modalities for support. Our approach better reflects real-world challenges and contributes a novel perspective to FSL.

\section{Additional Experiments}
\label{apendix:exp}

\paragraph{Hyperparameter Selection} \text{ }

We conducted a hyperparameter analysis to examine the impact of the number of latent domains $d$ on our model's performance using the Sketchy dataset under the All-way 1-shot and 5-shot settings. Figure~\ref{fig:latentdomain} illustrates how varying $d$ influences the accuracy across different modalities: mixed modality data, RGB data, and sketch data, with results averaged over multiple trials.

For \textbf{mixed modality data} (blue line), the accuracy remains relatively stable as $d$ increases, with slight improvements observed up to $d = 128$. Beyond this point, performance plateaus, indicating diminishing returns from increasing the number of latent domains further. The \textbf{RGB data} (green line) shows consistently high accuracy in the 1-shot setting, with minor improvements up to $d = 64$ in the 5-shot setting, after which performance stabilizes. The \textbf{sketch data} (red line) exhibits more variability, especially in the 5-shot setting, where accuracy decreases at intermediate values of $d$ but recovers and improves as $d$ approaches $128$. This suggests that the model's ability to capture sketch-specific features is sensitive to the number of latent domains, particularly when fewer domains are considered.

We selected $d = 128$ as the optimal number of latent domains for several reasons. First, at $d = 128$, the performance across all modalities is near its peak, which is crucial for our CFSL task, which relies on effectively handling multiple modalities. Second, the larger domain size helps stabilize the variability observed in sketch data, especially in the 5-shot setting, leading to more robust performance across both 1-shot and 5-shot scenarios. Lastly, while larger values of $d$ (e.g., $d = 256$) do not offer significant performance gains, they introduce additional computational overhead without clear benefits. Thus, $d = 128$ strikes a balance between performance and efficiency, enabling the model to effectively capture latent shared concepts across modalities.

Notably, the effect of increasing $d$ is less pronounced in low $k$-shot settings, particularly for smaller values of $k$. This diminished effect may be due to the limited number of latent domains being insufficient to capture the complex variations in the data when only a few examples per class are available. With small $k$, the model has less data to inform the latent space, making it challenging to effectively utilize a larger number of latent domains. Conversely, too many latent domains relative to the limited data can lead to overfitting or poor generalization. This highlights the importance of carefully selecting the number of latent domains in relation to the available data and the specific characteristics of each modality to optimize performance in few-shot learning scenarios.

\begin{table*}[t]
    \centering
	\caption{The details of the proposed GTL framework architectures. BS is short for batchsize, BN is short for BatchNorm1d. d determines the number of latent domains. }
        \label{tab:model_details}
	\setlength{\tabcolsep}{8pt}
	\begin{tabular}{lll}
	\toprule
	Module & Description & Dimenssions \\ 
        \midrule
        Encoder	 \\
        Input: visual representation $\mathbf{x}$  &  & BS $\times$ 1280 \\
        Dense &  \makecell[l]{256 neurons, with \\ BN, ReLU, Dropout } & BS $\times$ 256 \\
        Dense ($\mu$) &  \makecell[l]{mean of posterior \\ ($N_c$+$N_m$) neurons} & BS $\times$ ($N_c$+$N_m$)  \\
        Dense ($\sigma$)  &  \makecell[l]{variance  of posterior \\ ($N_c$+$N_m$) neurons} &  BS $\times$ ($N_c$+$N_m$)  \\
        Reparameterization & Sampling & $\mathbf{\hat{z}}_c$ ($N_c$) + $\mathbf{\hat{z}}_m$ ($N_m$)\\
        \midrule
        Disturbance encoder \\
        Input: visual representation $\mathbf{x}$ & & BS $\times$ 256 \\ 
        Gate  & Learnable gating function &  BS $\times$ d \\ 
        Dense & d * $N_m$ neurons & BS $\times$ d $\times$ $N_m$ \\
        Combination & Element-wise weighted sum & BS $\times$ $N_m$ \\
        Additional Aggregator Input: latent $\mathbf{\hat{z}}_m$ & & BS $\times$ $N_m$ \\
        Dense & Aggregation, $N_m$ neurons &  $\mathbf{\hat{z}}^{'}_m$($N_m$) \\

        \midrule
        Decoder
        Input: Concat ( $\mathbf{\hat{z}}_c$, $\mathbf{\hat{z}}^{'}_m$) & & BS $\times$ ($N_c$+$N_m$) \\
        Dense &  \makecell[l]{256 neurons, with \\ BN, ReLU, Dropout } & BS $\times$ 256 \\
        Dense &  \makecell[l]{1280 neurons} & BS $\times$ 1280 \\

        \midrule
        Classifier \\
        Input: latent intrinsic concept $\mathbf{\hat{z}}_c$ & & BS $\times$ $N_c$ \\
        Dense & 1280 neurons & BS $\times$ 1280 \\
        Dense & Classification output & BS $\times$ Class Number \\
        \bottomrule
	\end{tabular}
\end{table*}

\section{Network Architectures}
\label{apendix:network}

Table 4 provides a comprehensive overview of our GTL framework's architecture. We empirically set the intrinsic concept dimensionality $N_c$ as 128 and the modality-specific disturbance dimensionality  $N_m$ as 64 based on preliminary experiments that balanced model expressiveness and computational efficiency. The number of latent domains $d$ is set to 128, aligning with our hyperparameter analysis, indicating optimal performance at this value.

\textbf{Random Seed.} To ensure the reproducibility of our experiments, we set the random seed to 0 for all runs.

\end{document}